# An Open-Source Reproducible Chess Robot for Human-Robot Interaction Research

Renchi Zhang, Joost de Winter*, Dimitra Dodou, Harleigh Seyffert, Yke Bauke Eisma

*Faculty of Mechanical Engineering, Delft University of Technology, Delft, Netherlands*

* Correspondence: j.c.f.dewinter@tudelft.nl

**Abstract**
Recent advancements in AI have accelerated the evolution of versatile robot designs. Chess provides a standardized environment for evaluating the impact of robot behavior on human behavior. This article presents an open-source chess robot for human-robot interaction research, specifically focusing on verbal and non-verbal interactions. The OpenChessRobot recognizes chess pieces using computer vision, executes moves, and interacts with the human player through voice and robotic gestures. We detail the software design, provide quantitative evaluations of the efficacy of the robot, and offer a guide for its reproducibility. An online survey examining people's views of the robot in three possible scenarios was conducted with 597 participants. The robot received the highest ratings in the robotics education and the chess coach scenarios, while the home entertainment scenario received the lowest scores. The code is accessible on GitHub: https://github.com/renchizhhhh/OpenChessRobot

## 1    Introduction

Robots are becoming increasingly common across a variety of traditionally human-controlled domains. Examples range from automated mowers that maintain community lawns to robots on assembly lines and in agricultural settings. Recent advancements in AI have created new opportunities for intelligent sensing, reasoning, and acting by robots. In particular, the rapid development of large language models, such as ChatGPT and vision-language models, has lowered the barrier to human-to-robot communication by transforming text and images into interpretable actions or vice versa.

As technology advances, robots will likely attain greater capabilities and be able to tackle tasks previously within the exclusive realm of human expertise. This ongoing evolution may also lead to closer and more productive interactions between humans and robots. At the same time, integrating different AI-based robotic components remains a challenge, and the human-robot interaction (HRI) field lags in terms of endorsing reproducibility principles (Gunes et al., 2022). Encouraging transparent and reproducible research, therefore, remains an ongoing task.

The use of chess as a testbed for evaluating the effect of technology on human perception and behavior dates back to the 18th-century chess automaton Mechanical Turk (Standage, 2002). Furthermore, chess has played an important role in advancing the field of AI, starting with Claude

Shannon's chess-playing algorithm (Shannon, 1950) to the success of IBM's Deep Blue (Campbell et al., 2002) and DeepMind's self-play learning algorithm (Silver et al., 2018). In this paper, we incorporate AI algorithms into the design of a chess-playing robot to be used for studying HRI. HRI research may benefit from a chess-based setup because the game of chess provides a controlled, rule-based environment in which the impact of robots on human players can be precisely measured.

HRI-oriented studies with chess robots have typically used them to imitate human behaviors and assess their resulting impact on the human opponent. Pereira et al. (2008) and Leite et al. (2013) used the social robot iCat to play chess with children. This robot relies on an electronic chessboard as input and emits emotional responses and verbal utterances, guided by an emotion model. Sajó et al. (2011) developed Turk-2, a multimodal chess robot with human-like communication skills, while LC et al. (2021) explored human-robot and robot-robot-human interaction using artistic intervention, where expressive robot arms played chess and embodied distinct personalities.

When an electronic chessboard is not used, a camera is needed to determine if a move has been made and, if so, which move it was. A common solution is a monocular top-view camera, which has the advantage of not experiencing perspective-induced occlusion, making it relatively easy to identify any changes in the chess position (Golz & Biesenbach, 2015; Kołosowski et al., 2020; Larregay et al., 2018; Luqman & Zaffar, 2016; Mac et al., 2023; Srivatsan et al., 2020; Thanh Tra et al., 2018). An exception to this concept is the Gambit chess robot by Matuszek et al. (2011), which does not require a top-down view but uses a stereo camera instead. This robot is unique because it classifies the pieces by type and color, whereas other robots identify and track move-by-move changes from the starting position (e.g., Golz & Biesenbach, 2015; Kołosowski et al., 2020; Larregay et al., 2018; Siraj, 2017).

Other research focuses on the development of computer-vision methods for the detection of the chessboard, and the subsequent classification of the pieces on it, without always linking these computer-vision methods to a chess robot (Christie et al., 2017; Czyżewski et al., 2020; Ding, 2016; Koray & Sümer, 2016; Ranasinghe et al., 2023; Schwenk & Yuan, 2015; Xie et al., 2018). The detection of the chessboard typically involves line- or edge-detection techniques (Chen & Wang, 2019; Czyżewski et al., 2020; Srivatsan et al., 2020; Wölflein & Arandjelović, 2021; Xie et al., 2018), while the classification of chess pieces commonly uses convolutional neural networks (CNN) (Mallasén Quintana et al., 2020; Shin et al., 2023; Wölflein & Arandjelović, 2021). Challenges in both cases lie in testing for robustness under various conditions. A common technique to improve classification performance is to verify if the move/position is legal or plausible according to a chess engine. With such methods, it is possible to let the algorithm home in on the most probable classification outcome (e.g., Czyżewski et al., 2020; Mallasén Quintana et al., 2020; Ranasinghe et al., 2023). Currently, a state-of-the-art method is that of Wölflein and Arandjelović (2021), which used a diverse set of neural network models and achieved a per-square classification accuracy of 99.8%.

Our contribution centers on a software platform that others can reproduce using open-source robotics middleware. The chosen robot arm is widely available in research, while other hardware components can be substituted if required. The OpenChessRobot can detect pieces using computer vision and execute moves. For verbal interaction, we rely on large language models. Specifically, by connecting ChatGPT to the chess engine output, our robot interprets strategies and explains them to humans in natural language. Non-verbally, the robot is able to express its evaluation of the game by either nodding or shaking its end-effector. Supplementary Material Part

A provides a comparison of the OpenChessRobot with other chess robots designed for HRI studies.

This paper first presents the hardware and software design of the chess robot. It then provides an outline of its core modules, which include Perception, Analysis & Evaluation, Motion Planning & Execution, and Interaction. Module-specific experiments for performance evaluation are presented as well. Next, the paper reports findings from two human evaluations: an interview with an expert chess player who played two games with the robot, and an online study of 597 participants, which investigated perceptions of a talking robot equipped with AI in three different scenarios. Finally, the paper concludes with a discussion of the robot's limitations and outlines directions for future research in HRI.

## 2    Hardware and software architecture

Our chess-playing robot (Figure 1A) consists of the following hardware components (Figure 1B):
- a Franka Emika Panda robot arm (Franka Emika, 2020) equipped with a Franka Hand and a customized 3D-printed robot gripper. The robot arm has 7 degrees of freedom and is a widely used collaborative robot, suited for safe human interaction. It operates on firmware version 5.4.0
- a ZED2 StereoLabs camera (StereoLabs, 2020)
- a keyboard
- an external microphone and a speaker, connected to the PC
- an NVIDIA Jetson Nano (NVIDIA, 2019)
- a Linux PC with Ubuntu 20.04, running a real-time kernel. It is equipped with an Intel I7-8700K processor and an NVIDIA RTX 2080 graphics card.

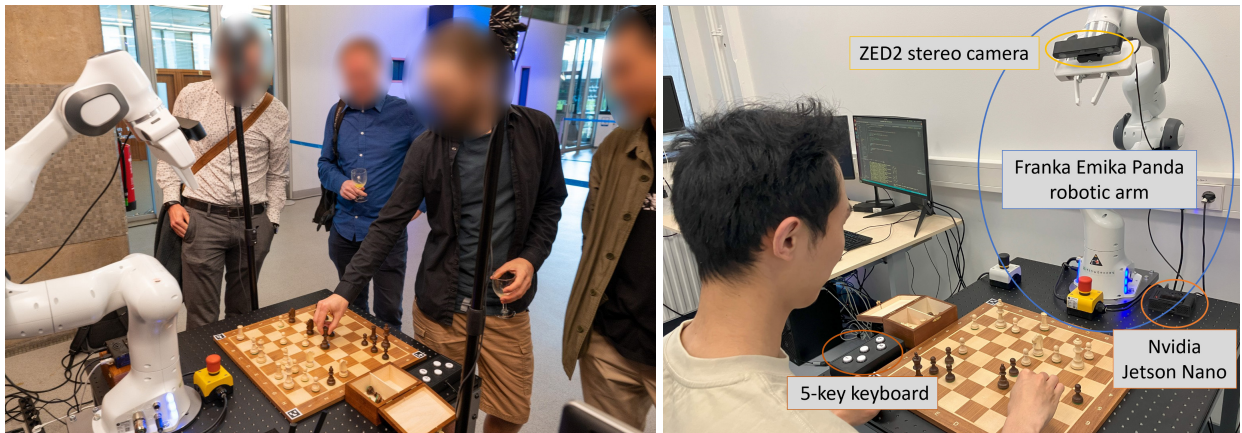

*Figure 1.* The system at a glance. (A) The robot at a demonstration event during which attendees could interact with it. (B) The complete system, comprising a Franka Emika Panda robotic arm, a ZED2 stereo camera, and an NVIDIA Jetson Nano computing board.

Our setup includes a number of accessories. These comprise a chessboard (48 × 48 cm), a set of chess pieces (model: Staunton No. 4; ranging in height from 4 to 7.8 cm and in width from 2 to 4 cm), and four printed ArUco markers. Considering the robot arm's range, the chessboard should be positioned so that the farthest rank remains within 65 cm of the robot. In Figure 1B, the distance between the closest edge of the board and the robot is 16 cm.

The robot arm, the Jetson Nano, and the Linux PC connect to a local network via a router. Communication between the arm and the control PC is realized through the Franka Control

Interface (Franka Robotics GmbH, 2023) by integrating the Libfranka (0.9.2) library for low-level control and the franka_ros package for the Robot Operating System (ROS) (Quigley et al., 2009). The ZED2 camera is attached to the Franka Hand via a custom mount, and it interfaces with the NVIDIA Jetson Nano through a USB connection. The Jetson Nano streams the camera view to the PC through ZED SDK v3.5. Considering that our setup uses images from a single camera, the ZED2 camera and Jetson Nano can be substituted with more economical monocular alternatives.

The software architecture of the OpenChessRobot is built upon ROS Noetic (Open Robotics, 2020), which offers common robotics data formats and message-passing among software modules. Figure 2 provides an overview of the software architecture, divided into four modules: Perception, Analysis & Evaluation, Motion Planning & Execution, and Interaction.

The Perception module uses the ZED2 camera SDK to capture the chess game images and relies on a neural network-based chess identifier to translate the game images into text descriptions. The latter identifier requires PyTorch and CUDA runtime. The Analysis & Evaluation module feeds the chess game annotation to a chess engine in order to get predicted moves and their corresponding scores. In the Motion Planning & Execution module, the predicted move, accompanied by the 3D chessboard localization results from the Perception module, is used by the MoveIt! motion planner (Coleman et al., 2014) to plan and execute chess moves.

The Interaction module takes the human inputs and manages the OpenChessRobot's verbal and non-verbal feedback. For the verbal feedback, a prompt wrapper, which combines the user request and outputs from the Analysis & Evaluation module, serves as a client of OpenAI's ChatGPT API service and generates the responses to talk with the human.

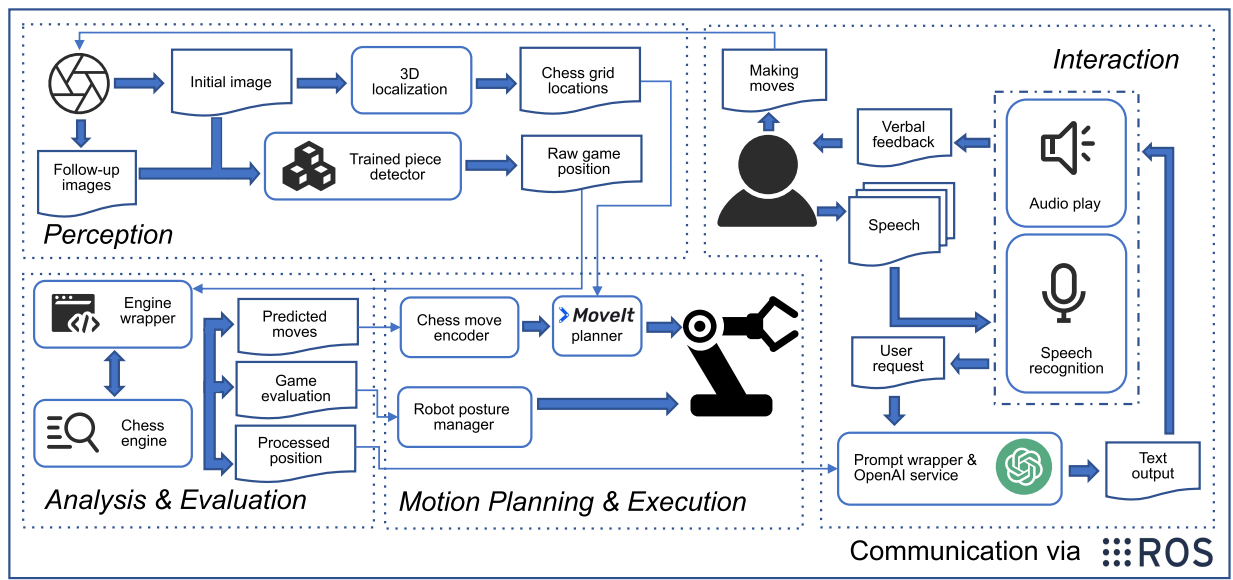

*Figure 2.* The software architecture of four modules and inter-module communication. Within the Perception module, the camera images are used to determine the 3D positions of the squares on the chessboard and identify individual pieces. The identified game position is analyzed through the chess engine wrapper, yielding the best next move and an evaluation of the current board position. The OpenChessRobot then enacts the chess move. Additionally, the robot is capable of responding through posture and speech.

## 3   Modules

The Perception module uses computer vision to identify the chessboard and the pieces; it can distinguish between occupied and empty squares and recognize pieces and their color. The perceived game position is sent to the Analysis & Evaluation module, which interfaces with a chess engine to determine the best move. The Perception module also calculates the chess squares' real-world locations, which are sent to the motion planner for planning the robot's end-effector trajectory to execute chess moves provided by the Analysis & Evaluation module.

The OpenChessRobot runs a data collection pipeline (orange box in Figure 3) and a human-robot gameplay pipeline (yellow box in Figure 3). The data collection pipeline is used to collect real-world data of chess pieces for retraining the Perception module and adapting to a new chessboard. The basic gameplay pipeline allows the robot to play the chess game with a human from any game position.

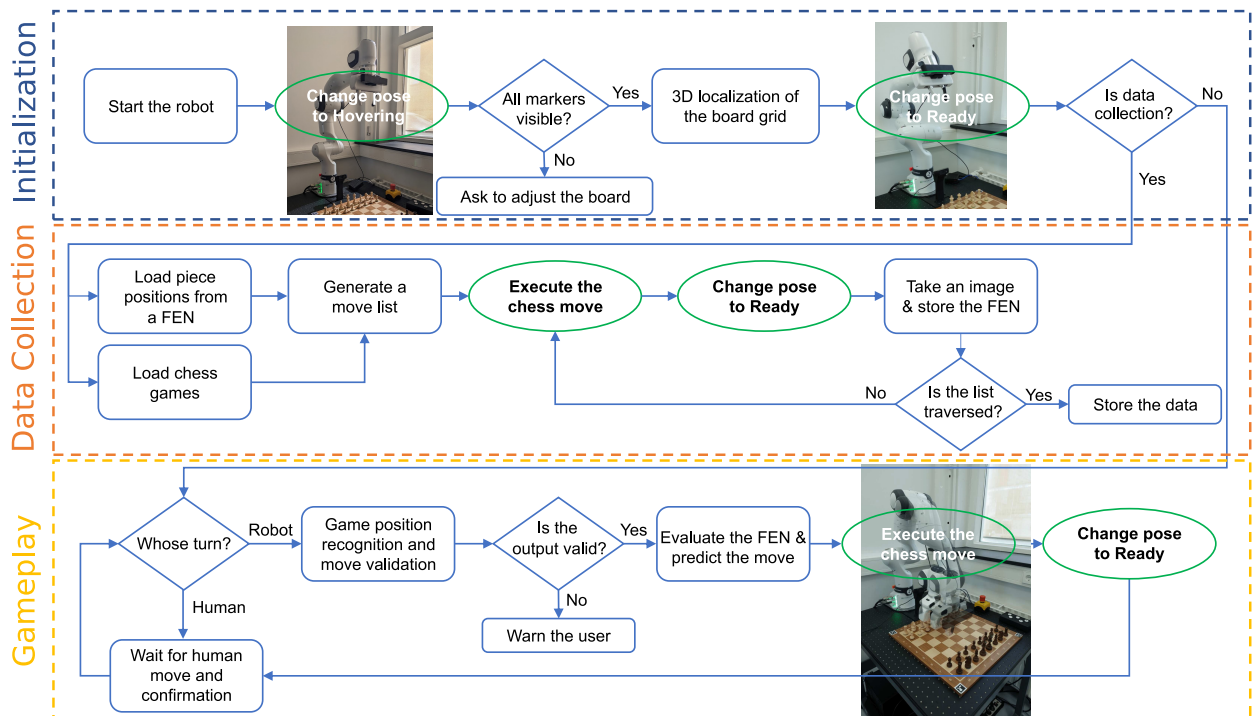

*Figure 3.* Flowchart of the OpenChessRobot execution using the core modules. The robot uses computer vision to recognize the chessboard and pieces, engages with a chess engine to determine moves, and executes chess moves for gameplay or data gathering. The chart splits into two key workflows: one for collecting data to improve the robot's perception of different chessboards and chess pieces, and another for playing chess with a human.

### 3.1   Perception

The ability to perceive the chess pieces is a key function of the chess robot. While the commonly used top-down view resolves the issue of piece occlusion, its practicality is limited by its narrow perspective on the pieces. Adopting lower camera angles and dynamic camera positions to observe the game supports the robot's natural interaction with the human player (see Figure 1). Additionally, our approach uses RGB images captured by a single camera instead of point clouds (Matuszek et al., 2011), improving the simplicity and reproducibility of our setup.

Figure 4 shows the Perception module of the OpenChessRobot. The module consists of two distinct classifiers, one for occupancy and the other for piece classification.

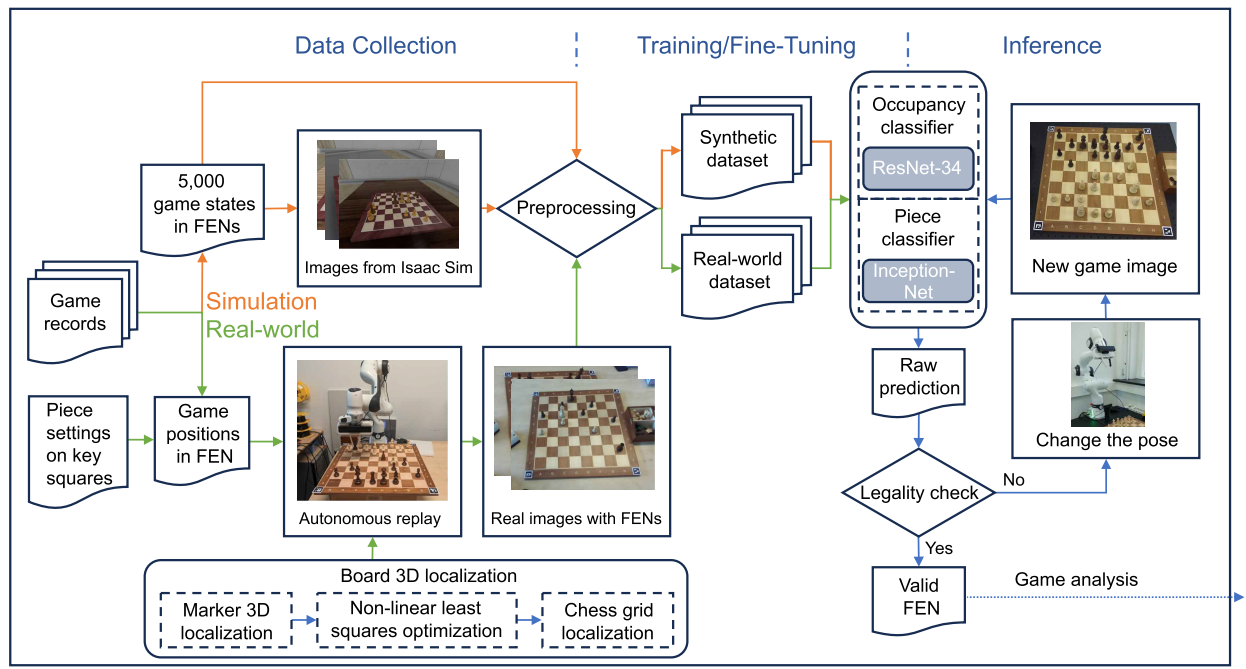

*Figure 4.* The Perception module of the OpenChessRobot relies on two distinct classifiers, one for occupancy and the other for piece classification. To train these classifiers, we synthesized chess positions in the NVIDIA Isaac Simulator (Makoviychuk et al., 2021) based on previous chess games and assigned ground truth labels, including the game positions represented in Forsyth–Edwards Notation (FEN) and pixel coordinates of board corners. In the real-world setup, the robot collects game images from the predefined camera pose (Figure 3) after autonomously placing pieces on key squares and replaying given games. These real-world images are then used to fine-tune the classifiers that were initially trained on synthetic datasets.

### 3.1.1 3D chessboard localization

To manipulate the chess pieces, their 3D coordinates in the real world need to be acquired. Four ArUco markers are used to indicate the 3D position of the chessboard. An image capturing these markers, along with the marker length and the camera's intrinsic and extrinsic parameters and distortion coefficients, allows for the computation of the marker-to-camera translation and rotation using OpenCV.

The 3D board localization is activated when the OpenChessRobot is initialized. The robot moves to its Hovering pose to observe the chessboard from above (blue box in Figure 3). Although the robot typically operates with low camera positions during the game, a vertical camera angle is used for 3D chessboard localization to reduce errors caused by image distortion. Then, a grid corresponding to the chess squares is formed based on the actual square size, and a non-linear least squares algorithm is used to optimize the grid corner positions to closely match the marker positions.

### 3.1.2 Chess piece detection

In the preceding section, we explained how the 3D coordinates of the squares are estimated. The next task is to identify the chess pieces on each square.

The piece detection model is an extension of a CNN-based model from Wölflein and Arandjelović (2021). We complemented their method by using a different synthetic dataset as well as a real-world dataset to fine-tune the pretrained models in order to effectively handle our real-world chess set.

To synthesize images of chess pieces on a board, we varied backgrounds, lighting conditions, camera poses, and piece locations. A total of 5,000 game positions, randomly selected from games played by grandmaster Bobby Fischer, were used. We synthesized images of selected chess games using the NVIDIA Isaac Simulator (Makoviychuk et al., 2021). Ground truth labels were created, consisting of the game positions represented in Forsyth-Edwards Notation (FEN) and the positions of the board corners. Following a similar approach to previous works (Mallasén Quintana et al., 2020; Matuszek et al., 2011; Wölflein & Arandjelović, 2021), the images were cropped into sub-images, each corresponding to a square on the chessboard. Two categories, "empty" or "occupied" were assigned for occupancy classification, while 12 categories (representing the six types of chess pieces in both black and white) were designated for piece classification.

To train the occupancy and piece classifiers with the synthetic data, 80% of the cropped images were used as the training set, while the remaining 20% was evenly split between a test set and a validation set. A ResNet-34 (He et al., 2016) and an InceptionV3 (Szegedy et al., 2016) were trained independently on their respective training datasets. This is illustrated in the left portion of Figure 4.

To adapt to real-world chess, a piece-square dataset was created by manually iterating over all the chess pieces on the board. During the iteration, the game positions were recorded both in FEN notation and in images taken at 1080p resolution from the perspective of the robot in the Ready pose (Figure 3). A number of training datasets to fine-tune the piece classifier were extracted from the piece-square dataset using sub-images of chess pieces on key squares. The trained models were evaluated on the remaining portion of the piece-square dataset. Additionally, to evaluate the classifiers' performance in games, a game dataset was collected by the robot autonomously replaying the provided chess games. Two games played by the grandmaster Michael Adams were used, with the game positions recorded both in FEN and in images, following the same procedure used when collecting the piece-square set. Dataset creation, training settings, and evaluation are explained in more detail in Section 5.1 and Supplementary Material Part B.

The preceding chess detection model considers only the position of the chess game. It does not verify the legality of the observed game position and does not possess knowledge of the previous game position. Such information can potentially improve the accuracy of the Perception module. Our legality check consists of two steps: first, it verifies if the game position conflicts with the chess rules, such as detecting no king or too many pawns. If the game position is invalid, it requests a new image (using the same camera pose but with potentially better camera exposure) and reanalyzes it. The robot can also change the camera angle to improve perception. More specifically, in case the predicted game states are detected as illegal, the robot slightly moves to its side in order to capture an image from a different perspective to improve the occupancy and piece classification.

In the second step of the legality check, the robot examines if the recognized chess move is among the legal moves from the previous game position. When the inferred game position and chess move prove to be legitimate, the robot adopts the inference and proceeds with gameplay. If not, the robot will halt its operation and wait for manual correction by the operator.

## 3.2 Analysis & evaluation

The processed chess FEN is forwarded to a chess engine wrapper, which uses the Universal Chess Interface (UCI) protocol for interaction with standard engines. Our chess engine wrapper builds upon an existing open-source Python wrapper for Stockfish (Zhelyabuzhsky, 2022). Our system integrates Stockfish 15 (Stockfish, 2022) as the default engine, using a total of 10 CPU threads. The chess engine assigns scores to candidate moves, which are monotonically related to the player's win rate, should the engine play against an equally strong opponent.

## 3.3 Motion planning & execution

For the planning of the OpenChessRobot's motions, we rely on the MoveIt! motion planner (Coleman et al., 2014). To determine the joint configurations, an inverse kinematics solver named IKFast is used (MoveIt, 2021). This solver calculates suitable robot joint angles based on the specified end-effector coordinates relative to the robot's base.

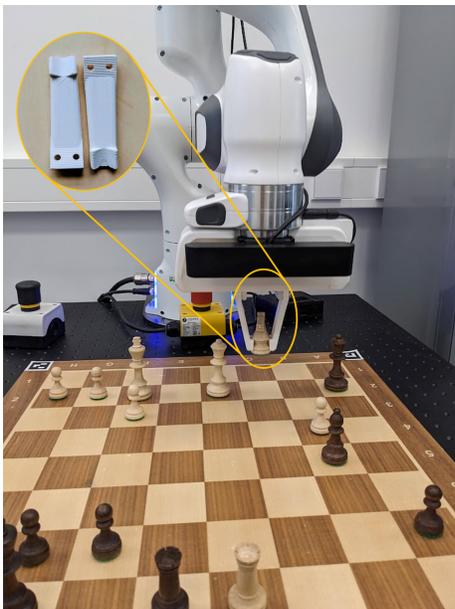

*Figure 5.* The customized 3D-printed gripper. This gripper has two fingers that are tolerant to deviations of chess pieces if they are not in the center of a square.

The robot arm is equipped with a customized gripper (Figure 5), designed to secure the underside of the piece, which is cylindrical in shape. The gripper is designed to be tolerant to minor deviations between the anticipated pickup positions (square centers) and the actual positions. Consequently, the human (or robotic) chess player is not bound to place a piece exactly at the center of a square.

There are two basic robot motions: a sliding motion (Figure 6A) and a jumping motion (Figure 6B). If the path between the start and end positions of a movement is unobstructed, the robot grasps the piece and pushes it to the destination square. When the path is obstructed, the robot raises the piece to leap over other pieces. Special movements like capture, castling, and en passant are manually programmed as combinations of the two basic motions.

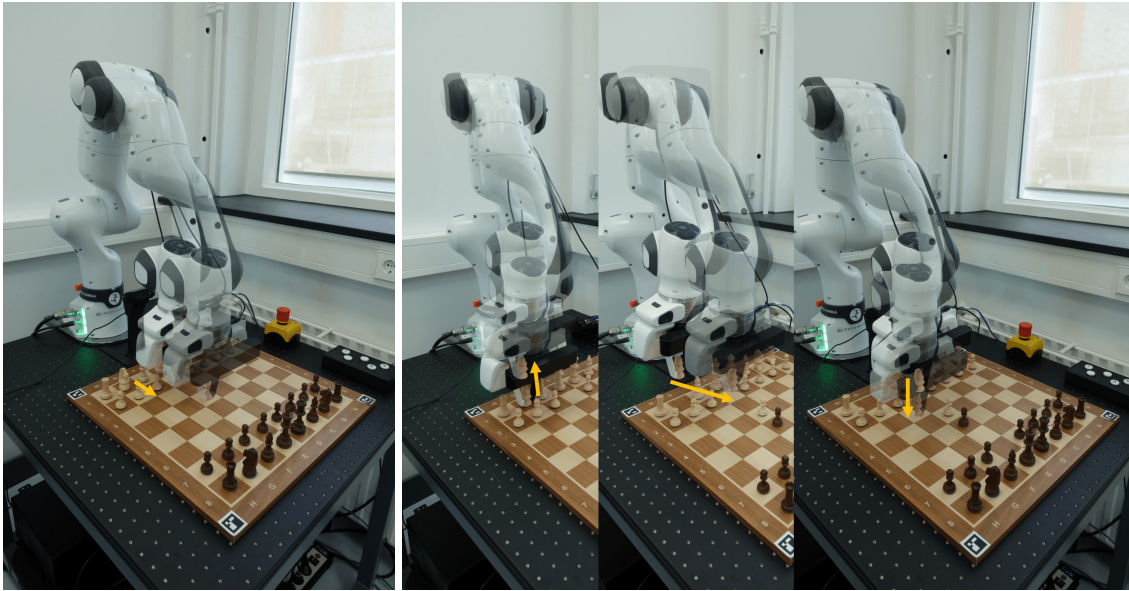

*Figure 6*. The robot is executing planned trajectories to move chess pieces. (A) The robot slides a pawn from the E2 square to the E4 square. (B) The robot lifts the knight on G1 to jump over other pieces on its way.

A chess move for the robot to execute is provided by the Analysis & Evaluation module in an encoded text format, e.g., g1f310000. The first four characters indicate the chess move. Additional information for a move is stored in the remaining characters. Specifically, the fifth to ninth characters indicate whether the move is a jump, capture, castling, en passant, or promotion move. With the encoded string, we create waypoints, using the 3D locations of the start square, the destination square, as well as intermediate states for special motions. The intermediate states are generated when the robot needs to lift pieces first rather than slide them on the board. The motion planner receives these waypoints and plans the trajectory of the end-effector using Rapidly-exploring Random Trees (RRTConnect; Kuffner & LaValle, 2000), with a 3D box surrounding the chessboard as a workspace constraint. After trajectory planning, velocity and acceleration limits are enforced on the planned trajectory (Kunz & Stilman, 2013).

## 4    Interactive modules

As the chess robot is designed for HRI research, providing various modalities for human-robot communication is important. We implemented an interactive gameplay pipeline based on the core modules (Figure 7). Specifically, depending on the human's chess performance, the OpenChessRobot can move its end-effector to express its evaluation, and it can provide verbal information regarding move qualities.

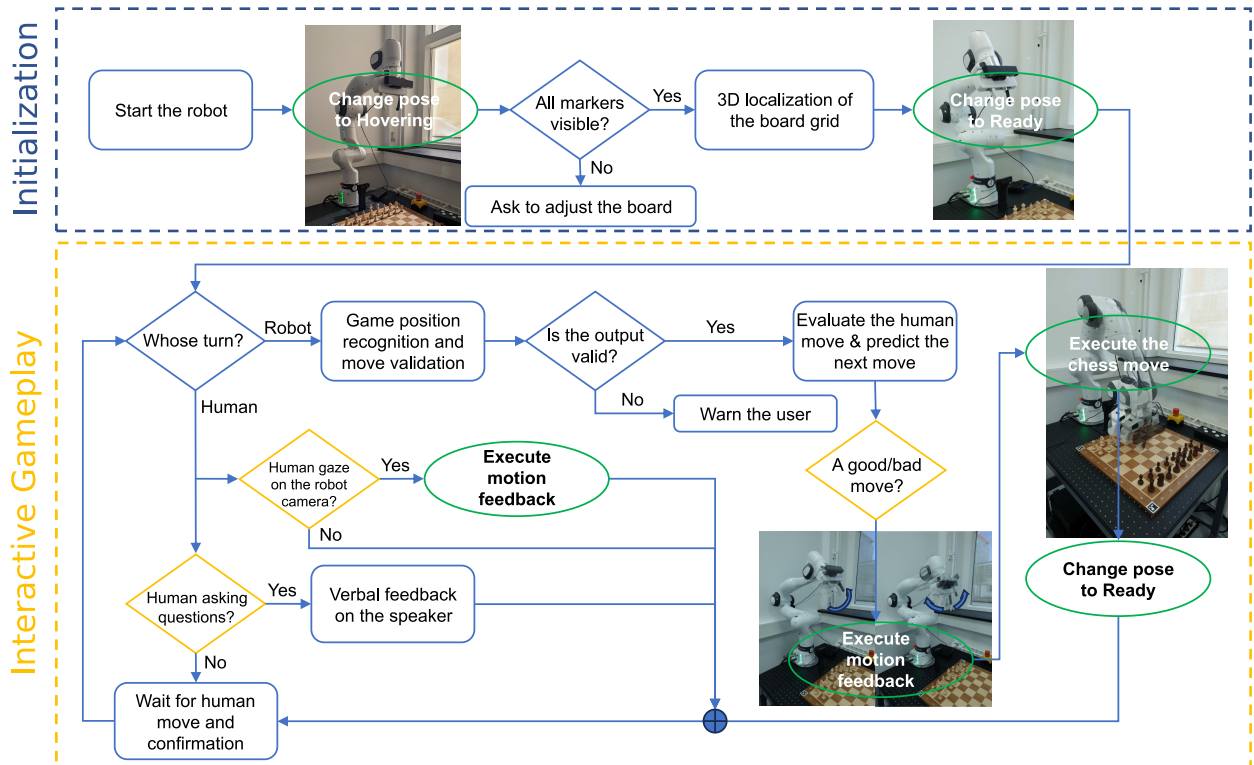

*Figure 7.* Flowchart of the OpenChessRobot execution using the core modules and Interactive module. As an extension of the core modules, the robot has an interactive gameplay pipeline. It evaluates human moves using predefined criteria and provides feedback by adjusting its posture. The system is also programmed to interpret human behaviors, such as gazing or asking questions, to provide a responsive and engaging experience.

### 4.1 Verbal interaction

In typical chess games, verbal communication does not hold a central role, although some professional players have explained the rationale and inner thoughts of chess moves in "Banter Blitz". Additionally, chess trainers commonly use verbal feedback to train their students. OpenAI's ChatGPT has proven capable of generating human-like dialogues. However, ChatGPT is unable to properly play chess on its own (Kuo et al., 2023). Our proposed solution relies on combining the analysis from the chess engine with ChatGPT without model fine-tuning.

Depending on the human player's request, the OpenChessRobot can explain the last move or the next move. The process is depicted in Figure 8. Capturing the user's request is achieved through voice recognition. To trigger the voice interaction, the robot should be in the Ready pose, waiting for the user's move. The player must vocalize specific keywords of the phrases, such as "explain the (last) move" or "analyze/predict the (next) move".

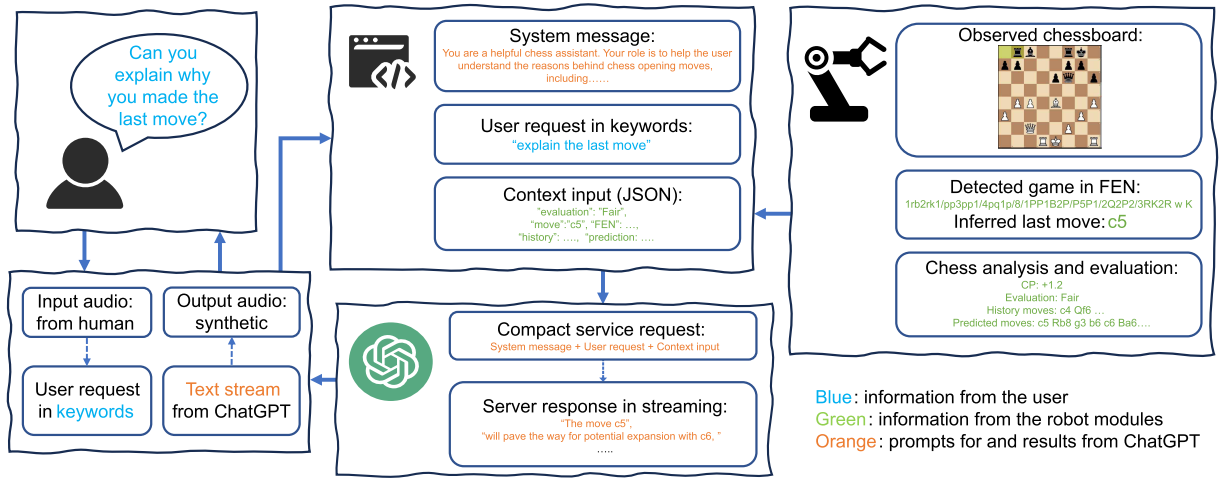

*Figure 8.* Verbal interaction. In this diagram, the user asks for an explanation of the last move. Next, a prompt consisting of a system message, the user request, and the processed game position is provided to the ChatGPT API to generate verbal feedback.

When explaining the last move, the move history and the current game FEN are sent to the Analysis & Evaluation module. Alternatively, when the user seeks advice on subsequent moves, the system uses the chess engine to predict several top-ranked moves, along with their game continuations. A wrapper integrates global instructions and the chess game information into a prompt ready for submission to the ChatGPT API (using model GPT-4-0613). The global instructions, which remain consistent across requests, are embedded in a system message (Table 1).

In response to the prompt, the ChatGPT API is requested to provide an analysis within the specified chess context in a tutor-like tone. To reduce the latency in vocalizing text responses from the server while maintaining a natural flow in synthetic voice, the feedback from the server, streaming in words/characters, is organized into short sentences, which are stored in a queue and played sequentially.

Table 1. *Prompt and ChatGPT output to explain the last move (bishop c4)*

| Prompt | |
|---|---|
| System message | You are a helpful chess assistant. Your role is to help the user understand the reasons behind chess opening moves, including their benefits, weaknesses, and overall strategic concepts. There are some instructions on how to respond to the user. Instruction 1: You should always start your response with a short sentence or few words to acknowledge that you've received the user's message and are considering it. Instruction 2: The user prefers concise explanations, so please keep your responses brief. Instruction 3: Your response should be informative and fluid, avoiding bullet points or index numbers. Instruction 4: When analyzing a board, mention one key reason for a move. Try to use the given history moves and predicted future moves. The input will include a user question, a chess FEN for the current game, the move to make, predicted future moves and a move history in a dictionary format under the keys 'question', 'fen', 'move', 'evaluation', 'future' and 'history'. Input: |
| User message | {'question':'can you explain?', 'fen':'r1bqkbnr/pppp1ppp/2n5/4p3/4P3/5N2/PPPP1PPP/RNBQKB1R w KQkq - 2 3', |
11

| |
|---|
| 'move':'Bc4', 'evaluation':'Excellent', \ 'future':'Bc4 Bc5 d3 Nf6 O-O d6 c3 O-O h3 h6 ',\ 'history': '1. e4 e5 2. Nf3 Nc6'} |
| **GPT-4-0613** |
| Sure, I can assist with that. The move Bc4 in this position is known as the Italian Game. The aim of this move is to control the center quickly with your pawn and knight and then put your bishop on its most dangerous square. After this, both white and black usually castle and then start a pawn storm against the opponent's king. In this case, future moves predict a traditional development of pieces with both sides castling kingside. The main goal is to have a solid position and try to exploit tiny advantages. |

## 4.2 Non-verbal interaction

In human-human communication, non-verbal cues play an important role. In professional chess games, verbal exchanges are often sparse. Apart from the moves played, participants may rely on behavioral cues to understand their opponents.

The posture feedback consists of basic nodding and shaking motions executed by the robot's hand (Figure 9). These gestures are triggered based on the chess engine's evaluation. Specifically, the robot enacts nodding or shaking gestures when the reduction in win probability due to the latest move exceeds a predefined threshold.

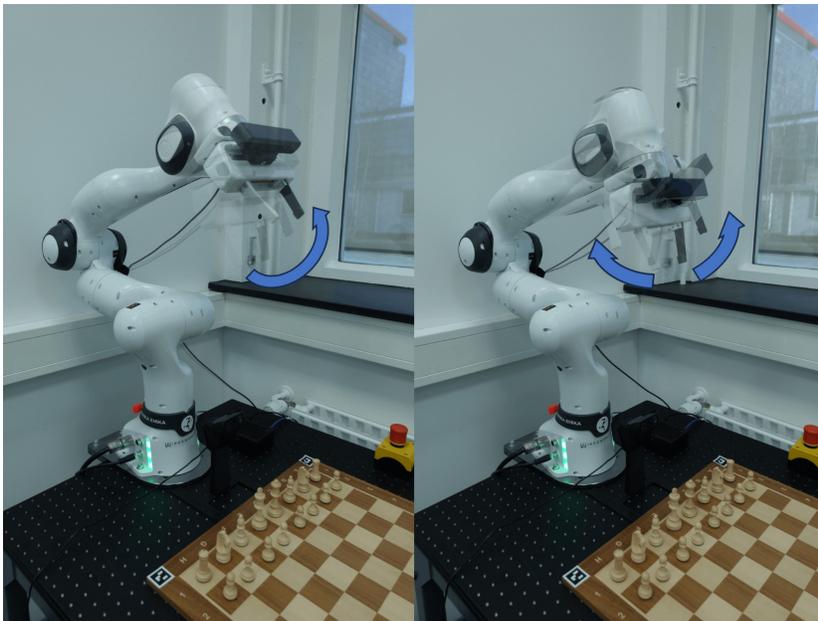

*Figure 9.* Posture feedback of the robot nodding and shaking its end-effector after a chess move was evaluated as good or bad, respectively.

## 5 Technical evaluation

We performed the following assessments of the chess robot system: (1) Perception module: accuracies of the piece classifiers; (2) Motion Planning & Execution module: success rates of grasping a piece from different sections of each square.

## 5.1 Chess perception

For the piece classifier of the chess perception module, several training settings were tested, using two sets of key squares, different components of the training dataset, and varying training lengths. The challenge was to identify a training setting that minimizes the number of training samples while achieving good generalization performance.

The two sets of key squares were:
- 3 × 3 squares, which include intersections between the key files A, E, and H and the key ranks 1, 4, and 8.
- 4 × 4 squares, which contain intersections between the key files A, E, and D, H and the key ranks 1, 4, 5, and 8.

The piece-square dataset contains three subsets: default, rotated, and visually shifted:
- The default dataset (**D**) contains sub-images of 768 unique piece-square pairs (64 squares × 12 pieces), with each pair occurring at least once.
- The rotated dataset (**R**) includes sub-images of four types of pieces: white king, white knight, black king, and black knight. Kings were rotated twice by 45 degrees (64 squares × 2 rotations), while knights were rotated four times (64 squares × 4 rotations) to create different shapes due to their rotational asymmetry (see Figure 10 right). Sixty-four samples of each rotated piece were randomly selected, forming the final rotated dataset.
- The visually shifted dataset (**S**) contains sub-images with cropping windows shifted in four directions (up, down, left, right). The displacement in each direction was ¼ square in length. For each shift direction, all 768 unique piece-square pairs were extracted. The **S** dataset serves as a test set with increased noise.

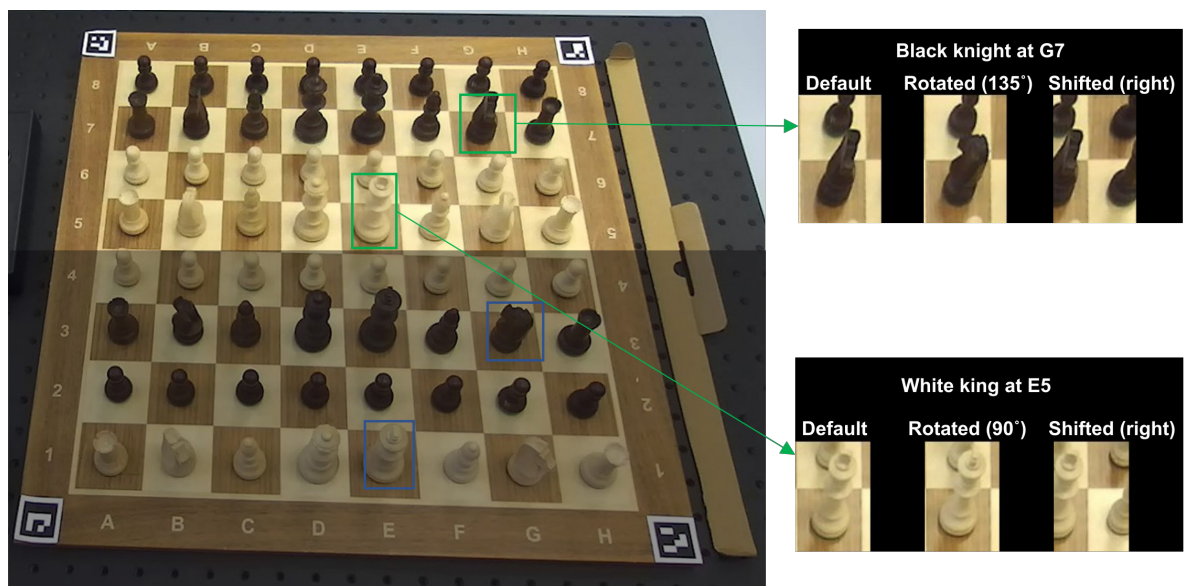

*Figure 10.* The raw images to create the piece-square dataset. The upper half image is taken without any rotations, and the lower half image is taken with 135-degree piece rotations.

Additionally, the data recorded by the OpenChessRobot while replaying the games was used as another test set (**G**) to evaluate the piece classifiers trained with piece images on the key squares. Only unique piece-square pairs in each game are extracted, making up the **G** dataset.

Different compositions of training and test datasets were extracted from these three datasets:
- The training set (**d**) contained sub-images of each piece on the key squares. The remaining samples in the **D** dataset were divided into a validation set and a test set in a 20:80 ratio.

- The training set (r) was generated from the R dataset using pieces on the key squares rotated by 90 degrees.
- Similarly, the training set (**rr**) contains all pieces on the key squares with all the rotations from the **R** dataset. The corresponding test sets and validation sets of **r** and **rr** were created in the same manner as the **D** dataset.

The fine-tuning was conducted in two phases: in the first phase, the weights of most layers of the model were frozen, and only the last layer (head) was trained over epochs with a learning rate of 0.001. In the second phase, all the model's weights were trained with a learning rate of 0.0001. The Adam optimizer was used in both phases.

Following the aforementioned training conditions, the piece classifier models were evaluated on their corresponding test sets, as well as on the **S** and **G** datasets. The evaluation results are detailed in Table 2. When comparing the two selections of key squares based on the model accuracies on the **G** dataset, it is evident that using 4 × 4 squares provides more samples for fine-tuning and results in good accuracies with better robustness. Additionally, including the rotated samples in the training set also helps improve performance, but it requires a longer training duration to be effective. Results on the **S** dataset show that using 4 × 4 squares provides better robustness against the noise caused by piece displacements compared to using 3 × 3 squares. While longer training durations and adding rotations to the training set yield some improvements, these improvements are not substantial.

Regarding the occupancy classifier, it was found to achieve 100% accuracy on the **G** dataset after retraining for 50 epochs using piece sub-images at the 4 × 4 squares and all the adjacent empty squares.

Table 2. *Piece classifier accuracies in different training conditions. The best accuracies on the S and G datasets are underlined.*

| Test set | Key squares | 3 × 3 | | | | | | | | | 4 × 4 | | | | | | | | |
|---|---|---|---|---|---|---|---|---|---|---|---|---|---|---|---|---|---|---|---|
| | Epoch | 500+500 | | | 1000+500 | | | 1000+1000 | | | 500+500 | | | 1000+500 | | | 1000+1000 | | |
| | Training set | d | d+r | d+rr | d | d+r | d+rr | d | d+r | d+rr | d | d+r | d+rr | d | d+r | d+rr | d | d+r | d+rr |
| D-d | | 0.9924 | | | 0.9943 | | | 0.9943 | | | 0.9978 | | | 1.0000 | | | 1.0000 | | |
| D+R-d-r | | | 0.9794 | | | 0.9835 | | | 0.9863 | | | 0.9939 | | | 0.9939 | | | 0.9954 | |
| D+R-d-rr | | | | 0.9915 | | | 0.9887 | | | 0.9930 | | | 0.9984 | | | 0.9922 | | | 1.0000 |
| S | | 0.9762 | 0.9704 | 0.9684 | 0.9704 | 0.9691 | 0.9759 | 0.9652 | 0.9688 | 0.9740 | 0.9837 | 0.9831 | 0.9844 | 0.9834 | 0.9883 | <u>0.9886</u> | 0.9814 | <u>0.9886</u> | 0.9844 |
| G | | 0.9932 | <u>1.0000</u> | 0.9932 | 0.9865 | 0.9662 | <u>1.0000</u> | 0.9932 | 0.9865 | <u>1.0000</u> | 0.9865 | <u>1.0000</u> | <u>1.0000</u> | <u>1.0000</u> | <u>1.0000</u> | <u>1.0000</u> | <u>1.0000</u> | <u>1.0000</u> | <u>1.0000</u> |

### 5.2 Grasping of pieces

An experiment was performed in which the OpenChessRobot was tasked with grasping various pieces positioned on different squares of the chessboard, repeated multiple times. To reduce positional discrepancies, a compensation of 1 cm along the *y*-axis and 0.55 cm along the *x*-axis (Figure 11) was applied as the system error.

Specifically, the evaluation focused on testing the grasp performance using the largest piece, the king (4 cm in width, 7.8 cm in height), and the smallest piece, the pawn (2 cm in width, 4 cm in height). These pieces were positioned at each of the four corner squares (A1, A8, H1, H8) and physically shifted from the edges of the squares to their centers (depicted in Figure 11). Two shifting distances were used: (1) placing the piece tangent to the square's edge or (2) 0.625 cm away from the edge, corresponding to 1/8 the length of a square. Each piece in each square position was tested 10 times.

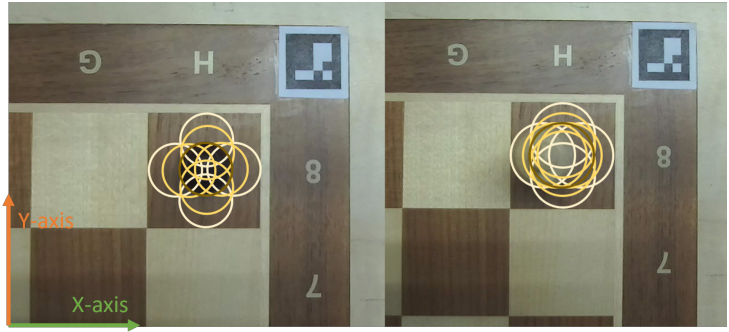

*Figure 11.* The positions by physically shifting the piece at square H8. In the left image, the black pawn is in the center without shifting. In the right image, the white king shifts to the left edge of the square. The brown circle shows the center position. The white circles indicate the first shifting distance (to the edge), and the yellow circles indicate the second shifting distance (intermediate).

Figure 12 shows the success rates for the different independent variables of the grasping experiment. A distinction is made between three categories: (1) *accurate grasping:* a correct grasp in 10 out of 10 trials, (2) *remedied grasping:* a correct grasp in 8 or 9 out of 10 trials (usually correct grasping occurred here in such a way that the piece slid into the grasper's teeth), and (3) *missed grasping*: a correct grasp in 0 to 7 out of 10 trials. Table 3 shows the corresponding results in numerical form. A trend can be noticed, whereby the grasping was least successful when the piece was positioned toward the edge of the square or toward the 'up' direction, i.e., in the positive direction of the *y*-axis (see Figure 11), positioned away from the base of the robot arm. It is hypothesized that this effect stems from image projection and coordinate transformation errors in the 3D chessboard localization.

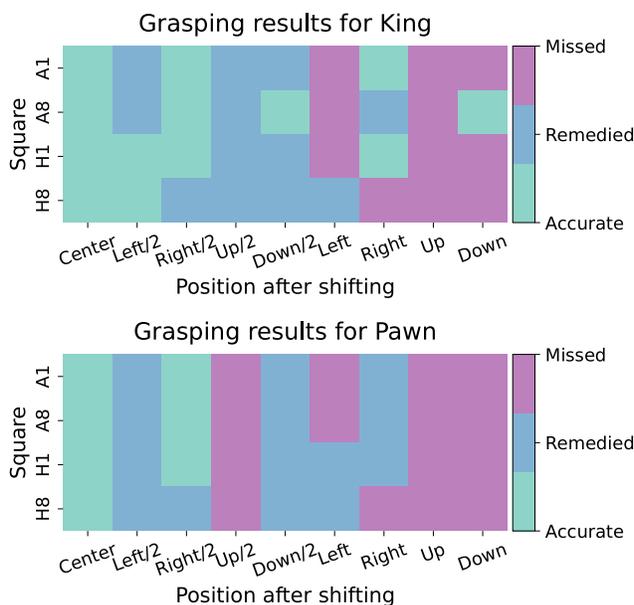

*Figure 12.* Results from the grasping experiment.

Table 3. *Piece grasping results. The percentages represent the number of trials that were marked as 'accurate grasping' or 'remedied grasping'.*

| Piece | Overall | | Shift Distance | | Shift Direction | | | |
|---|---|---|---|---|---|---|---|---|
| | Without Shifting ($n$ = 4) | With Shifting ($n$ = 32) | Halfway ($n$ = 16) | Edge ($n$ = 16) | Left ($n$ = 8) | Right ($n$ = 8) | Up ($n$ = 8) | Down ($n$ = 8) |
| Pawn | 100% | 65% | 75% | 31% | 75% | 88% | 0% | 50% |
| King | 100% | 78% | 100% | 31% | 63% | 88% | 50% | 63% |

### 5.3 Performance in human-robot chess play

To assess the core components of the chess robot, a chess game between a human and the OpenChessRobot was conducted. The robot had control of the white pieces while the human played with the black pieces. To determine its next move, Stockfish 15, with a search depth of 20, was used.

The OpenChessRobot initiated the game from its Ready pose, executed a move, and subsequently returned to its Ready pose, awaiting the human player's move. The human player was prompted to press the middle keyboard button (see Figure 1B) upon completing their move. The manual button press was maintained to mirror traditional chess tournament practices and ensure clear separation of moves. This design also maintains safety and user control safeguards by providing an explicit signal before the robot begins its movement. The robot then captured an image, identified the human's move, and made its subsequent move. Two demo videos of human-robot chess play can be found in Supplementary Material Part C.

To ensure a fair match, the human player also had access to Stockfish 15 to determine their next best move. The game ended when a checkmate was achieved or a draw due to repetition. Throughout the experiment, we recorded the time taken by each robot module as it carried out its respective tasks.

The game ended in a draw after 97 moves, including 49 robot moves and 48 human moves. Out of these 49 robot moves, 7 involved capturing pieces, wherein the robot first removed the captured piece before executing its move, 6 moves required jumping pieces over others, resulting in longer execution times compared to moves made on unobstructed paths, and one was the castling. The other 35 moves followed the pick-and-slide approach.

The time taken for the OpenChessRobot to execute its moves is depicted in Figure 13A. On average, it took 7.33 seconds for the robot to complete a chess move, from its initial position to the resumption of that position after the move. Capture moves, on average, consumed 6.24 seconds. Figure 13B shows the time allocation for move detection, evaluation, and prediction. Move detection was typically completed within 1 second, with one outlier attributed to camera failure. Evaluating a single move with Stockfish took approximately 1 second. Finding the next move with Stockfish generally took approximately 5 seconds, unless Stockfish had already encountered the current human move during its previous search. In such cases, the search time was virtually zero, allowing the robot to execute the next move immediately. The relatively large computation time can be explained by the fact that Stockfish was allowed to search at a high depth, which is typically unnecessary for amateur-level play.

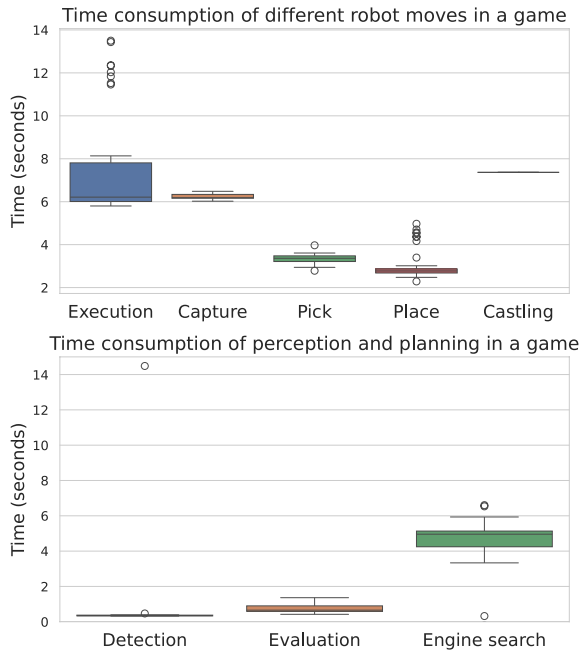

*Figure 13.* (A) Time consumption of different robot moves in the game. (B) Time consumption of move detection, Stockfish evaluating a move, and Stockfish searching for the next move.

## 6  Evaluation of people's views on the chess robot

To collect an expert chess player's opinion on the OpenChessRobot, we involved a FIDE Master to play against the robot in two games (Supplementary Material Part D). The interview conducted after playing two games revealed that the robot, while potentially useful for beginner-level practice, lacks human-like feedback or commentary from which experts might benefit. The player also indicated that the robot might be useful for entertainment purposes.

In addition to the evaluation by an expert player, we conducted an online study to investigate how the broader population perceives the OpenChessRobot and to gather opinions on the future uses of such embodied AI in general. Participants watched a demo video of the robot and completed a corresponding questionnaire.

Specifically, demo videos were shown featuring the robot in three scenarios:
- The scenario *Robotics Education* featured the robot explaining how it works. The 28-second demo video showed the robot describing its perception function and how it localizes the chessboard.
- In the *Chess Coach* scenario, the robot teaches a chess opening. The 36-second video demonstrates specific moves and highlights key squares on the chessboard.
- The scenario *Home Entertainment* represents a future household robot that intends to entertain people through conversation and chess play. The 42-second demo video showed the robot discussing a chess game from a movie (Campbell, 1997). The videos and corresponding transcripts can be found in Supplementary Material Part C.

The choice of these scenarios is partly based on the aforementioned experiences with the expert chess player and also represents how (chess) robots could potentially be used in the future. That is, chess robots do not necessarily have to provide chess lessons (*Chess Coach* scenario) but can also support education by explaining their own functionality and serving as a knowledge resource (*Robotics Education* scenario). Additionally, robots can be seen as a companion or interaction partner (*Home Entertainment* scenario).

## 6.1 Participants

We aimed to recruit 600 participants through the online research platform Prolific (Douglas et al., 2023; Peer et al., 2021) from six countries where English is an official or de facto primary language: Australia, Canada, Ireland, New Zealand, the United Kingdom, and the United States.

A questionnaire was launched on Prolific on Wednesday, February 12th, 2025, at 22:00 Central European Time. In Prolific, prospective participants were provided with a hyperlink to the survey platform Qualtrics to complete our questionnaire titled 'User perception of a talking chess robot'. The purpose of the study was described as "*to examine how people evaluate a talking chess robot, which is powered by artificial intelligence including ChatGPT*". Once the target of 600 active participants was reached, the survey was automatically closed. Ultimately, 597 responses were collected. Respondents gave informed consent via a dedicated questionnaire item. Each respondent received a reward of £1.50 for completing the estimated 10-minute questionnaire. The research was approved by the TU Delft Human Research Ethics Committee (Application Number 5109).

## 6.2 Experiment design

The online study followed a between-subjects design, with each participant presented with one of the three demo videos. All participants first answered questions about their familiarity with ChatGPT. After this, a short text was presented introducing one of the three demo videos, such as (*Chess coach* scenario): "*The following video shows a talking chess robot that uses artificial intelligence (including ChatGPT) to give a chess lesson. The robot explains a chess opening. Please turn on your sound.*"

Next, the demo video was played, followed by questions about people's acceptance (Q1), impression of the technical soundness of the robot (Q2), possible improvements (Q3), and opinions on the potential for such talking robots equipped with AI to replace existing technology (Q4) or human workers (Q5). The first three questions collected people's perceptions of the current OpenChessRobot, and the last two questions aimed to acquire their opinions on the future use of such talking robots equipped with AI. The five questions were as follows:

Q1. "*Do you think this robot could be used <in schools to explain to students how a robot works?> / <in schools to teach students chess?> / <at home to provide entertainment?>*" (Definitely not, Probably not, Neutral, Probably Yes, Definitely yes, I prefer not to respond)

Q2. "*Comment on the technical quality of the robot as shown in the video. Mention positive or negative aspects.*" (A response of at least 20 characters was required)

Q3. "*Do you have any recommendations about how to improve the robot?*" (A response of at least 20 characters was required)

Q4. "*Do you think that within 10 years, robots like this could serve as a replacement of existing screen-based <computers to help understand technical topics such as computer vision?> / <computers to improve chess skills?> / <forms of entertainment such as television?> Please elaborate on why or why not.*" (A response of at least 20 characters was required)

Q5. "*Would you welcome a future where robots that can talk, equipped with artificial intelligence (such as improved versions of ChatGPT), replace human workers?*" (Definitely not, Probably not,

Neutral, Probably Yes, Definitely yes, I prefer not to respond)

The remaining questions gathered demographic information (age, gender, and education) and assessed participants' attitudes toward and familiarity with chess, technology, and the movie used in the *Home Entertainment* scenario. The full questionnaire can be found in Supplementary Material Part E.

### 6.3 Data analysis

For the open-ended questions (Q2–Q4), we used ChatGPT o1 (OpenAI, 2024) to analyze the responses (see Supplementary Material Part F for the full prompts and raw outputs). Previous research shows the possibility of using large language models to analyze textual data from user studies (Tabone & De Winter, 2023). In addition, the reasoning model ChatGPT o1 has demonstrated its efficiency in textual content classification for social media analysis, sentiment analysis (Zhong et al., 2024) and categorical reasoning (Latif et al., 2024). We requested ChatGPT o1 to extract the three most mentioned positive and negative aspects from responses to Q2, the five most mentioned recommendations from responses to Q3, and the three most mentioned concerns from responses to Q4.

Additionally, for the open-ended question Q4, we used two reasoning models, ChatGPT o3-mini-high and Gemini 2.0 Flash Thinking, as text classifiers to classify the responses regarding people's opinions into *positive*, *neutral*, or *negative*. If the two classifiers produced different predictions, a human annotator manually determined the final label.

### 6.4 Results

A total of 195, 211, and 191 participants were randomly allocated to the *Robotics Education*, *Chess Coach*, and *Home Entertainment* scenarios, respectively. The median time to complete the survey was 6.48 min. The mean age of the participants was 36.5 years ($SD$ = 12.33, $n$ = 597). The gender distribution was: 302 females (50.6%), 286 males (47.9%), 8 '*other*' (1.3%), and 1 '*I prefer not to respond*'.

Figure 14 presents the distribution of responses for Q1, Q4, and Q5. For Q4, the two reasoning models classified the opinions identically for 528 participants, while a human annotator manually labeled the remaining 69 responses.

#### 6.4.1 Analysis of acceptance per scenario

Acceptance was the highest for the *Robotics Education* scenario (Figure 14A), with a mean of 3.96 on a scale of 1 ("*Definitely not*") to 5 ("*Definitely yes*") ($SD$ = 0.98, $n$ = 195) (Q1). According to the ChatGPT o1 analysis, the most mentioned positive aspects for this scenario were the "Advanced technical ability" in terms of computer vision and the "*Clear voice/explanation*" demonstrated by the robot (Q2). However, participants often found the technical language in this scenario rather hard for beginners or students to comprehend.

The acceptance of the *Chess Coach* scenario was the second highest, with a mean of 3.82 ($SD$ = 1.05, $n$ = 210, excluding one participant who chose '*I prefer not to respond*') (Q1). The most mentioned positive aspect specifically for this scenario according to ChatGPT o1's analysis of Q2 was that it was seen as an "*effective teaching approach*". However, participants also often mentioned the "*limited interactivity*" as a downside.

The acceptance of the Home Entertainment scenario was the lowest (mean = 3.59, $SD$ = 1.07, $n$ = 191), with only 15% of the participants answering "*Definitely yes*" to Q1. A similar pattern can be observed in the classified opinions of Q4, where a majority of participants (57%) expressed a negative expectation regarding the use of such a robot to replace screen-based entertainment like television (Figure 14B).

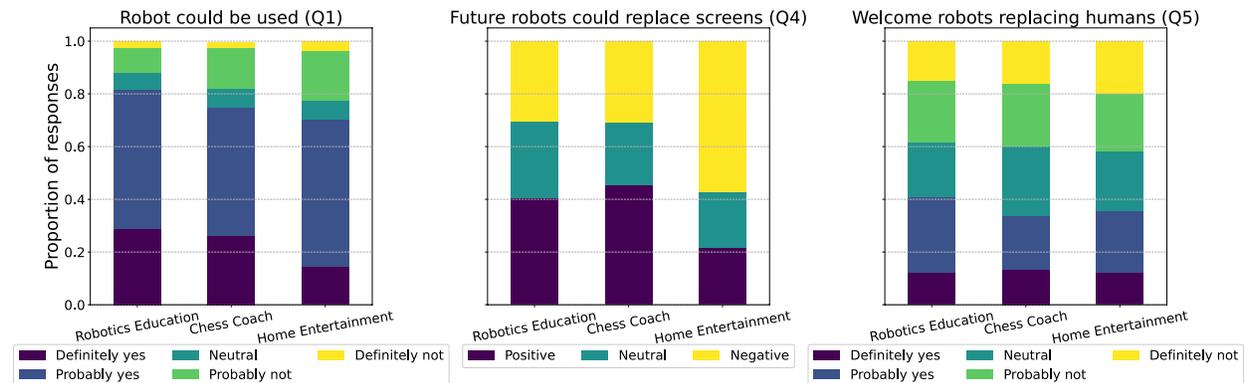

*Figure 14.* (A) The proportion of responses to Q1 "*Do you think this robot could be used …*", (B) Q4 "*Do you think that within 10 years, robots like this could serve as a replacement of existing screen-based … Please elaborate on why or why not.*" (answers classified into three categories), and (C) Q5 "*Would you welcome a future where robots that can talk, equipped with artificial intelligence (such as improved versions of ChatGPT), replace human workers?*".

### 6.4.2 Analysis of open-ended questions across the three scenarios

The summary results of the open-ended questions are shown in Table 4. Participants frequently praised the smooth and precise movements of the robot and its clear teaching style, but they criticized its robotic voice, mechanical noise, and slow operation. Common recommendations included adopting a more natural, human-like voice, simplifying the explanations, reducing the mechanical noise, speeding up performance, making the design more friendly, while keeping the robot cost-effective.

Table 4. *Summary of results from the open-ended questions analyzed by ChatGPT o1. Underlined items indicate aspects that were commonly mentioned across all three scenarios.*

| | **Theme** | **Robotics Education** | **Chess Coach** | **Home Entertainment** |
|---|---|---|---|---|
| Q2 | The three most mentioned positive and negative aspects | Positive:<br>• Smooth and precise movement<br>• Clear voice/explanation<br>• Advanced technical ability (vision, analysis)<br>Negative:<br>• Overly technical/Complex language<br>• Slow or lengthy process<br>• Robotic or distracting Voice/Noise | Positive:<br>• Smooth and precise movement<br>• Clear explanations<br>• Effective teaching approach<br>Negative:<br>• Robotic or unnatural voice<br>• Noisy or distracting movements<br>• Limited interactivity | Positive:<br>• Smooth and precise movement<br>• Intelligent or knowledgeable<br>• Entertaining or interactive<br>Negative:<br>• Loud or noisy<br>• Bulky or large<br>• Slow or boring |
| Q3 | The five most mentioned recommendations | • More natural, varied, or human-like voice<br>• Simpler, age-appropriate explanations<br>• Friendlier or more appealing design<br>• Reduced mechanical noise | • More natural, simpler speech<br>• Reduce mechanical noise<br>• Friendlier design and smaller size<br>• Greater interactivity and Q&A | • Reduce mechanical noise<br>• Use a more natural, expressive voice<br>• Make the design more compact<br>• Improve overall aesthetics/approachability<br>•    Enhance movement |

|   |   | • Faster or smoother operation | • Clearer movement cues | fluidity |
|---|---|---|---|---|
| Q4 | The three most mentioned concerns | • High cost and limited cost-effectiveness<br>• Acceptance and human interaction<br>• Incomplete or restricted functionality | • High cost<br>• Limited portability<br>• Reduced versatility | • Lack of variety and visual/storytelling depth<br>• Preference for passive entertainment<br>• Insufficient human/emotional element |

# 7 Discussion and conclusion

In this paper, we introduced an open-source cognitive robot designed for engaging in chess matches with humans. While multiple chess robots have been developed over the years, none have been made available as a reproducible platform, and many of them have not been specifically designed for conducting HRI studies. Our robot takes a unique approach by integrating robust robotic perception, evaluation of gameplay, and move execution with verbal and non-verbal interactions. Our development prioritized adaptability and reproducibility, to create an accessible platform for researchers and enthusiasts alike.

We focused on the game of chess because of its value as a controlled experimental environment where both human behavior and machine performance can be accurately measured. Chess also serves as a 'battleground' for Moravec's paradox, which states that while machines excel at computational tasks (such as playing chess), they traditionally struggle with tasks requiring human-like perception, motor control, and language processing (Moravec, 1988). Our research platform aims to transcend Moravec's paradox by integrating chess engine capabilities with innovations in human-like perception and the verbalization of moves using synthetic speech.

Furthermore, the integration of verbal and non-verbal interactions is included to increase the depth of engagement between humans and the robot. By using a large language model for verbal communication, we tried to create a solution that lowers the barrier of understanding the reasons behind robot chess moves. Our platform, in essence, revisits the conceptual foundation laid by the Mechanical Turk in the 18th century (Standage, 2002).

We invited an expert chess player to play with the platform and found that although it can be useful for beginners, it failed to provide human-like in-depth feedback from which expert-level players might benefit. Additionally, we conducted an online survey with 597 participants recruited from six countries to gather their views about the OpenChessRobot as well as talking robots equipped with AI in general. The online survey results revealed that participants favored the robot's use in educational contexts (where its advanced technical skills were appreciated) while expressing reservations about its role in home entertainment, primarily due to concerns over limited interactivity and its inability to replace traditional screen-based media. These findings motivate future research on HRI in educational settings using this robot.

One of the limitations of the OpenChessRobot, as also shown in Table 4, is that its movement is rather slow and noisy. Another limitation is that the ChatGPT-based explanations are still narrow in scope and quality. To improve the robot, we will focus on faster arm movements and the incorporation of advanced large language models that are specifically trained or fine-tuned on chess-related text databases (see also Feng et al., 2023). Here, there is a need for explainable AI, where the chess position, represented by a FEN notation, is translated into a verbal explanation that is not only correct in terms of the chess engine's evaluation (as our current robot already does) but also contains meaningful content about why a particular move is good or bad.

Moreover, although the results from the online study demonstrate promising aspects of the platform, any application beyond the research setting, such as education, coaching or entertainment, should be researched further. Finally, a limitation of the current paper is the lack of physical human-subject experiments among a large number of chess players. In the future, we intend to use this setup to study how AI-embodied robots influence people during interactions. This will involve the robot communicating with humans through emotional expressions and more natural/meaningful verbal interactions.

In conclusion, this paper presents the OpenChessRobot, an open-source, reproducible cognitive chess robot that integrates robust computer vision, chess engine evaluation, and both verbal and non-verbal interactions for HRI studies. A large language model is used to translate the evaluation from the chess engines into human-like speech. Supported by an international survey and expert gameplay, our findings show the robot's potential in educational and research contexts, while highlighting limitations in human-like interaction.

**Funding**
The author(s) declare that financial support was received for the research, authorship, and/or publication of this article. Specifically, this project has received funding from the Cohesion Project of the Faculty of Mechanical Engineering, Delft University of Technology.

**Data availability statement**
Models for chess recognition used in the OpenChessRobot software, material from the online study, images used to finetune and evaluate the models, raw data, and scripts for reproducing the analyses in the paper are available at a public data repository: https://doi.org/10.4121/1cb5bf64-468e-462a-a82e-c847d88a7a86

# Supplementary Material

## Part A. Comparison with existing chess robots

Table A1. *A comparison between the OpenChessRobot and other chess robots designed for human-robot interaction (HRI), evaluating aspects such as perception, manipulation, reproducibility, interactive features, and main purpose. The selected works explicitly mention the use of the robot in HRI studies or list "human-robot interaction" among their keywords.*

| Robot | Perception | | Manipulation | | Reproducibility | | Interactive Features | Main Purpose |
|---|---|---|---|---|---|---|---|---|
| | Method | Evaluation | Method | Evaluation | Material | Code/dataset | | |
| Pereira et al. (2008), iCat | Relying on E-board | Not applicable | Relying on E-board | Not applicable | iCat robot, E-board | Not available | Facial expressions | To study "the influence of embodiment in the enjoyment of a game" (p. 1253) |
| Sajó et al. (2011), Turk-2 | Via an on-top RGB camera | Not reported | Via a robot arm | Not reported | A camera, a customized robot arm, and a standard chess set | Not available | Facial expressions, speech of fixed terms | To present a chess robot that can "communicate with the opponent" (p. 493) |
| Matuszek et al. (2011), Gambit | Via a RGB camera | Piece Recognition: 93.22% | Via a robot arm | Success rate: 91.6% (in three games) | A camera, a customized robot arm, and a normal chess set | Not available | Not available | To present a chess robot "that can play physical board games against human opponents" (p. 4291) |
| LC et al. (2021), Power chess | Memorizing the game based on previous moves | Not applicable | Via a robot arm | Not reported | Two robot arms | Not available | Gesture expressions | To study the human perception of robot-to-robot gesture expressions in chess scenarios |
| Mac et al. (2023) | Via an on-top RGB camera | Not reported | Via a robot arm | Not reported | A camera, a Reactor X150 robot arm, and a standard chess set | Not available | Not available | To constructing a physically capable Chess-Playing system to compete against the player |
| OpenChessRobot | Via a RGB camera | Piece Recognition: 96.52%~100% | Via a robot arm | Success rate with a pawn: Center: 100% Halfway: 75% Edge: 31%; more results in the paper | A camera, a Franka Panda robot arm, a customized gripper, a normal chess set | Open-source code, 3D-print models, and dataset available. | Gesture expressions and generative speech for game analysis | To represent an open-source reproducible chess robot that can play physical chess games with humans, interpret the game position verbally, and provide feedback with gestures |

**Part B. Dataset creation**

A new dataset was collected by placing all the pieces (63 pieces total, one rook missing) on the board, and then taking photos from the default camera pose. The corresponding labels in FEN format were then created. After performing row movements, column movements, and rotations, each piece appeared at least once at each square on the board, with rotations. The procedure for creating the dataset is detailed below. The dataset and fine-turned models can be found at: https://doi.org/10.4121/1cb5bf64-468e-462a-a82e-c847d88a7a86

**Starting position**
The pieces were manually placed in the middle of the square. The camera pose was such that pieces often occluded each other. Each piece was placed so that its adjacent neighbors included both white and black pieces.

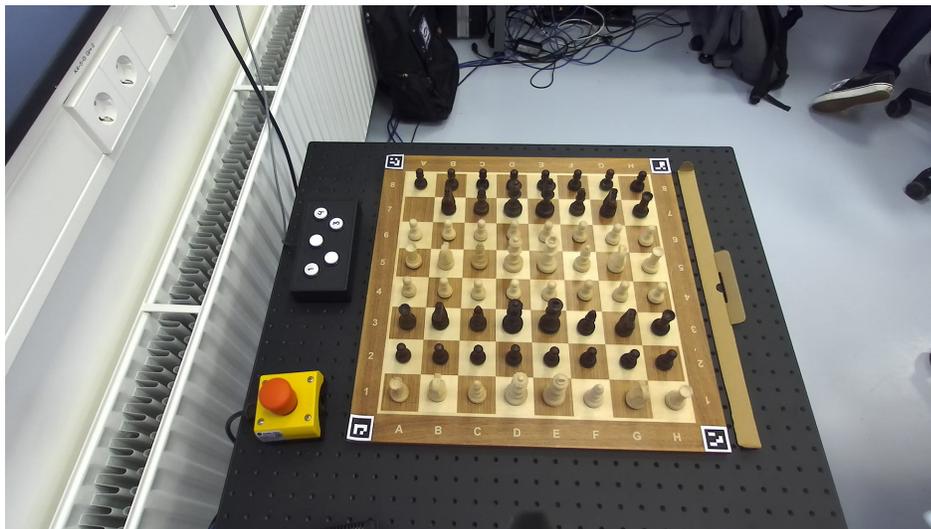

*Figure B1.* Starting position. Image in dataset: chess_image_data/raw/img/img_C0_R0_r0.png. Label in FEN: pppppppp/1nbqkbnr/PPPPPPPP/RNBQKBNR/PPPPPPPP/rnbqkbnr/pppppppp/RNBQKBNR

**Taking photos of the default and rotated piece positions**
At a game position, multiple photos were taken: one photo showed the default setting, and four more photos showed the pieces rotated. Four types of pieces were rotated because of their rotational asymmetry: white king, white knight, black king, and black knight. Kings were rotated twice, and knights were rotated four times to create different shapes of the pieces in the image. Each rotation was approximately 45 degrees.

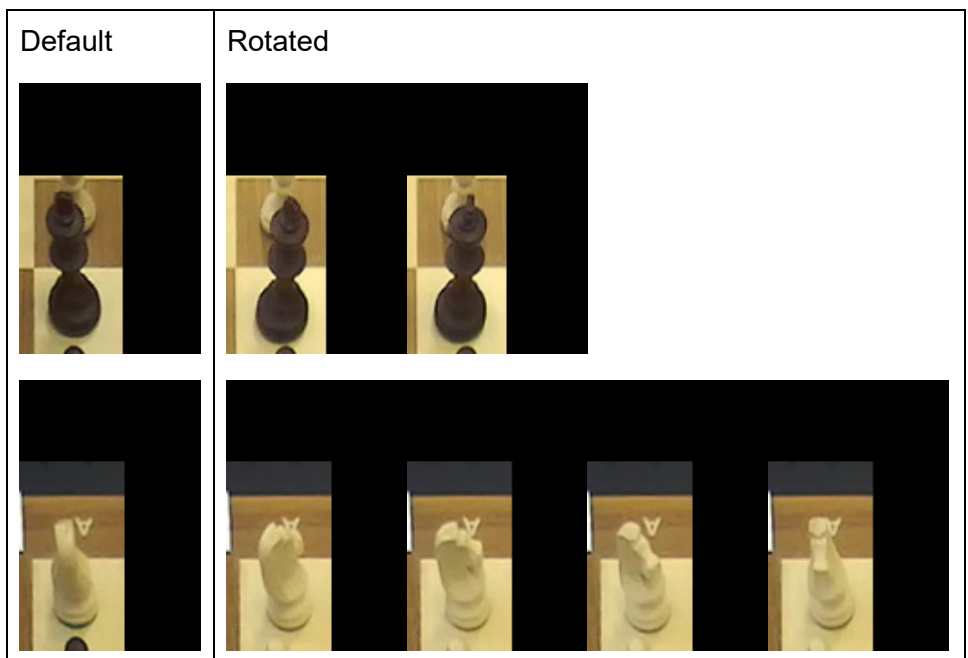
*Figure B2.* Illustration of piece rotation.

Following the above procedure, at each chess piece's position, five images were taken (unrotated, both the knight and king rotated twice, knight rotated twice more). Photos are named as: img_CX_RY_[r0, r1, r2, r3, r4].png

**Translating the pieces by row**

In the direction of increasing rank, the rows of pieces were moved to make a new chess piece position. Then five photos were taken at the new position with the same method mentioned above. The row translation was repeated three times (R0 > R1 > R2 > R3).

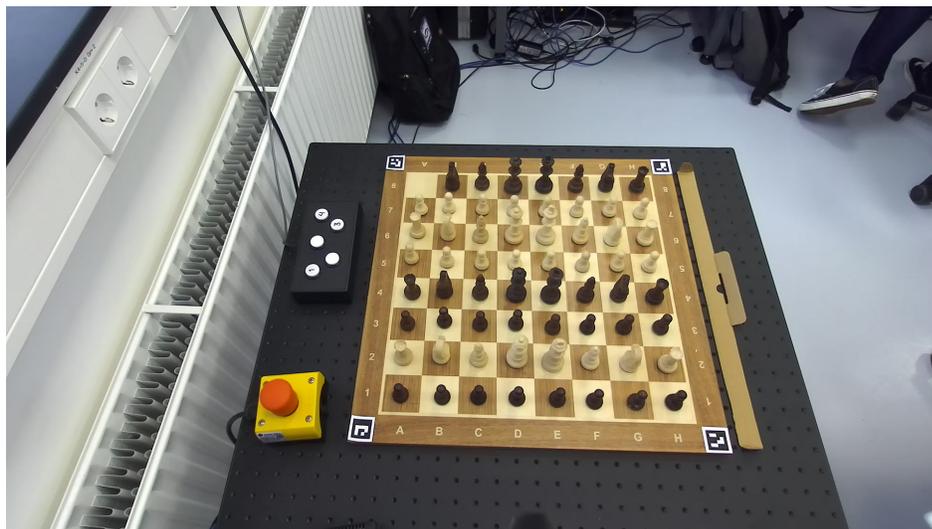
*Figure B3.* Starting position, after one upward row shift. Image in dataset:
chess_image_data/raw/img/img_C0_R1_r0.png. Label in FEN:
1nbqkbnr/PPPPPPPP/RNBQKBNR/PPPPPPPP/rnbqkbnr/pppppppp/RNBQKBNR/pppppppp

**Translating the pieces by column**
After translating the row three times, one additional row translation was realized (R3 > R0), in order to restore the R0 game position. After that, a column translation in the (A > H) direction was performed (C0 > C1), followed by taking photos at this new game position using the same method mentioned above.

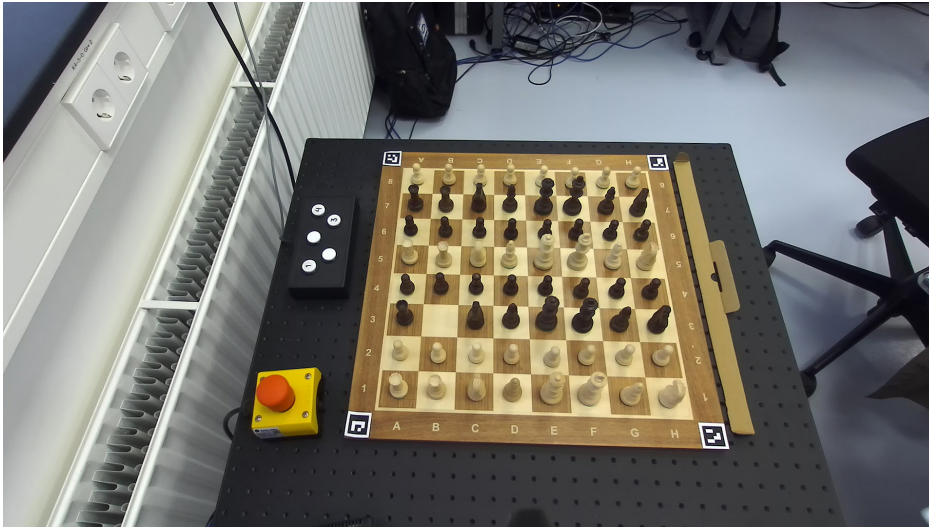
*Figure B4.* Starting position, after one rightward row translation. Image in dataset: chess_image_data/raw/img/img_C1_R0_r0.png. Label in FEN: PPPPPPPP/rrnbqkbn/pppppppp/RRNBQKBN/pppppppp/r1nbqkbn/PPPPPPPP/RRNBQKBN

**Creating the datasets**
A total of 160 raw images were collected (4 row movements × 8 column movements × 5 orientations), from which three datasets were created: default, rotated, and visually shifted.
The default dataset (**D**) contained 2048 sub-images of unrotated pieces extracted from the raw images, with every unique piece-square combination represented at least once.
The rotated dataset (**R**) contained 1280 sub-images of each rotated piece (black king, white king, black knight, and white knight) extracted from the raw images, with every unique piece-square-rotation combination represented at least once.

The visually shifted dataset (**S**) contained 8192 sub-images with cropping windows shifted in four directions (up, down, left, right). The displacement in each direction was ¼ of a square's length. For each shift direction, 2048 sub-images were extracted from the raw images without piece rotations, following the same procedure used to generate the dataset **D** but with displaced cropping windows. The visually shifted dataset is used only for testing, not for training.

**Training data selection and train-val-test split**
Several squares were selected as key squares for the board. Sub-images of pieces on these squares were used to train the model; this model was then tested on the remaining squares to show its generalization ability. Two sets of key squares were tested:

1. 3×3:
    key_file: A, E, H;
    key_rank: 1, 4, 8;
    training set: 9 × 12 sub-images;

2. 4×4:
   key_file: A, E, D, H;
   key_rank: 1, 4, 5, 8;
   training set: 16 × 12 sub-images;

The training dataset (d) from the default dataset was generated using all sub-images from these key squares. One sub-image was chosen randomly for each square-piece pair.
The training dataset (r) was generated from the rotation dataset using all sub-images rotated by 90 degrees. One sub-image was chosen randomly for each square-piece pair with the same rotation angle.

Another training dataset (rr) was generated from the rotation dataset using all sub-images rotated by both 45 degrees and 90 degrees. For black and white knights, one sub-image was chosen randomly for each square per rotation angle.

Then, sub-images of the remaining squares were split into a validation set and a test set using a ratio of 20:80. The test sets corresponding to datasets d, r, rr, respectively, are referred to as d, r, rr.

Before feeding the training set to train the model, some random image augmentation was applied to each sub-image. These parameters were set through trial and error.

Table B1. *Parameters for augmentation.*

| Attributes | Range |
| --- | --- |
| brightness/hue/saturation | [-20%, 20%] |
| contrast | [-30%, 30%] |
| scale | [0.8, 1.2] |
| translation | [-0.03, 0.1] |

Examples:

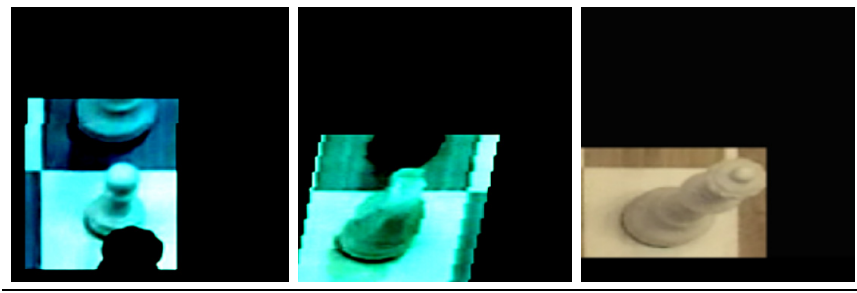

**Part C. Links to videos of the OpenChessRobot**

Video 1: Demonstrations of the robot's movements and game explanation. Human-robot gameplay with body posture feedback is also included. Link:
https://www.youtube.com/watch?v=RenXuiwX4Go

Video 2: A sample from an annotated video recording using the Tobii Glasses 3 eye-tracker from the perspective of a professional chess player. To protect the privacy of the participant, audio has been removed. Link: https://www.youtube.com/watch?v=j-b32ILQjtw

Video 3: The three videos used in the online questionnaire survey. Link:
https://www.youtube.com/shorts/Mj_rZVboH9s

Transcripts of video 3:
The *Robotics Education* scenario: *I determine the exact 3D position of the board first by detecting the four ArUco markers placed on it and calculate the chessboard's 3D grid. Then once I identify which square each piece is on through my occupancy and piece classification, I map that square to its corresponding 3D coordinates on the board.*

The *Chess Coach* scenario: *Great, I am playing e4 which is one of the most popular and aggressive openings for White, controlling the center and opening lines for my queen and bishop. Now if Black plays c5 they are introducing the Sicilian Defense. Instead of directly placing a pawn in the center (like with e5), Black plays c5 to exert influence on the d4 square.*

The *Home Entertainment* scenario: *This is the famous position from 2001: A Space Odyssey, where Frank Poole faced against the HAL 9000. In the movie Poole as White captured on a6 with the queen and walked right into a forced mate. But if we analyze carefully White might try something safer—like shifting the queen to b7 instead of immediately grabbing the pawn. Or pushing a rook to e1. What would you like me to try?*

**Part D. A case study with an expert chess player**

On May 25, 2023, during development, we invited a FIDE Master to play against the initial version of the OpenChessRobot. His feedback was used to improve the interactive features, leading to the current version of the OpenChessRobot.

Two games were played between the human and the robot. In the first game, the player played against a Stockfish chess engine. In the second game, the player experienced the verbal feedback on the game positions, while the robot used the chess engine Maia ELO 1900 (McIlroy-Young et al., 2020), which was trained on human games. The player wore Tobii Pro Glasses 3 during the games to record the gaze activity and was instructed to think aloud (a video can be found in Supplementary Material Part C). An experimenter prompted the player to explain strategies and thoughts. After the games, an interview was conducted to gather feedback on the player's experience with the robot, potential use cases, and suggestions for improvements.

The recordings were processed using AssemblyAI's Speech-to-Text API for speech recognition and speaker diarization. OpenAI's ChatGPT o1 model was used to translate the transcriptions from Dutch to English and to answer specific questions (see Table D1). Eventually, only the recording of the interview was analyzed, as most comments during the games were about the game strategy.

Table D1. *Questions to ChatGPT o1 and corresponding responses.*

**Question 1**: What is the playing experience of the chess player playing against the chess-playing robot? Provide concise answers in bullet points.

**Answer from ChatGPT o1**:
- The chess player did not find the robot intimidating and felt it resembled playing a regular human opponent primarily due to its strong moves.
- The novelty of a physical, tangible opponent added a different dynamic compared to playing chess online.
- The player expressed that faster, more natural robot movements would make the experience more enjoyable.
- The absence of meaningful feedback (e.g., identifying mistakes or strong moves) reduced the robot's value as a training tool.
- The player would find the robot more beneficial if it played and provided insights in a more human-like and instructive manner.

**Question 2**: What does the chess player think about the possible use cases and the unlikely use cases of this chess robot? Provide concise answers in bullet points.

**Answer from ChatGPT o1**:
- Likely Use Cases:
  - Practicing specific chess openings or positions.
  - Receiving instructive feedback on moves and tactics.
  - Serving as a training tool for beginners.
  - Providing an entertaining, physical alternative to playing online.
- Unlikely Use Cases:
  - Playing random casual games where no human-like or instructive element is added.
  - Replacing standard online engines for straightforward chess play.
  - Serious professional training relying solely on a "perfect" engine (like Stockfish).

> **Question 3**: What are the suggestions from the chess player to improve the robot? Provide concise answers in bullet points.
>
> **Answer from ChatGPT o1**:
> - Offer immediate feedback on moves, including identifying blunders or strong plays.
> - Employ a more human-like engine (e.g., the Maia engine) to create a realistic playing style.
> - Increase the speed and fluidity of the robot's physical movements.
> - Provide instructive commentary that explains why certain moves are good or bad.
> - Allow scenario-based training, such as practicing specific openings or positions.

From the interview, we learned that the player did not find the robot intimidating and recognized its potential use for beginners in practice or entertainment scenarios. However, he noted that the robot could be improved to play more like a human, both in terms of game strategy and motion speed. To improve the robot, he suggested it should provide direct feedback on move quality and move faster.

The eye-tracking data from the two games was processed into annotated videos, showing where the player was looking during the game. The gaze data provides insights into how the player anticipated the robot's next move and how often the player engaged in gaze interaction with the robot's physical body (see Figure D1). These gaze recordings offer valuable data for future HRI studies on this platform. A video clip of the eye-tracking data is included in Supplementary Material Part C (Video 2).

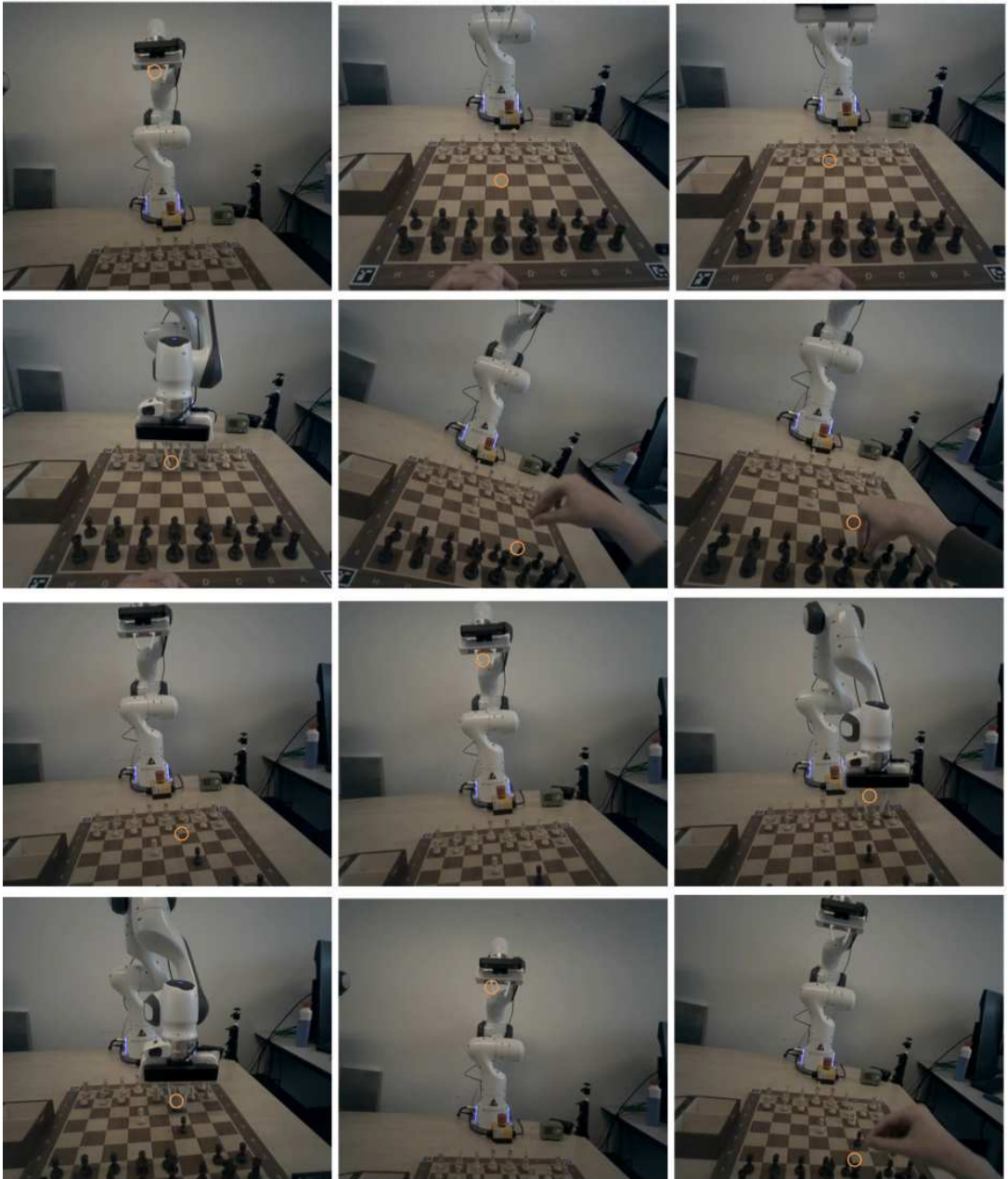

*Figure D1.* Annotated video with gaze overlay. Row 1 (left to right): The player looked at the robot and scanned the board for the robot's possible moves. Row 2 (left to right): The robot made its first move, followed by the player's first move. Row 3 (left to right): The player predicted the robot's next move and looked at the robot, waiting for its execution. Row 4 (left to right): The robot executed its move, the player gazed at the robot again, then made the second move.

**Part E. The full questionnaire**

| Category | Question |
|---|---|
| GPT familiarity | Q1. *"Have you ever heard of ChatGPT?"* (Yes, No, I prefer not to respond)<br>Q2. *"How often do you use ChatGPT?"* (Every day, 4-6 days a week, 1-3 days a week, About once every 2 weeks, About once a month, Less than once a month, Never, I prefer not to respond)<br>Q3. *"Do you use a paid version of ChatGPT (Plus or Pro)?"* (Yes, No, I prefer not to respond) |
| Video Demo | |
| Perception of the robot | Q4. (A/B/C) *"Do you think this robot could be used <in schools to explain to students how a robot works?> / <in schools to teach students chess?> / <at home to provide entertainment?>"* (Definitely not, Probably not, Neutral, Probably Yes, Definitely yes, I prefer not to respond)<br>Q5. *"Comment on the technical quality of the robot as shown in the video. Mention positive or negative aspects."* (A response of at least 20 characters was required)<br>Q6. *"Do you have any recommendations about how to improve the robot?"* (A response of at least 20 characters was required)<br>Q7. (A/B/C) *"Do you think that within 10 years, robots like this could serve as a replacement of existing screen-based <computers to help understand technical topics such as computer vision?> / <computers to improve chess skills?> / <forms of entertainment such as television?> Please elaborate on why or why not."* (A response of at least 20 characters was required)<br>Q8. *"Would you welcome a future where robots that can talk, equipped with artificial intelligence (such as improved versions of ChatGPT), replace human workers?"* (Definitely not, Probably not, Neutral, Probably Yes, Definitely yes, I prefer not to respond) |
| Demographics | Q9. Age, Q10. Gender, Q11. Education |
| Familiarity with the context of scenarios | Q12. *"Are you a technology enthusiast?"* (Not at all, Slightly, Moderately, Very, Extremely, I prefer not to respond)<br>Q13. *"Are you familiar with the movie 2001: A Space Odyssey?"* (Not at all, Slightly, Moderately, Very, Extremely, I prefer not to respond)<br>Q14. *"Did you ever play in an official FIDE-rated chess tournament?"* (No, Yes, I prefer not to respond)<br>Q15. *"Are you a chess enthusiast?"* (Not at all, Slightly, Moderately, Very, Extremely, I prefer not to respond) |

**Part F. Prompts for ChatGPT o1 and the raw outputs**

| | |
|---|---|
| Prompt to ChatGPT o1 to analyze Q5 | Based on the given answers to the question: "Comment on the technical quality of the robot as shown in the video. Mention positive or negative aspects.", summarize the most mentioned three positive and negative aspects in the answers. Make concise bullet points. Answers: |
| Output of Scenario A | Most Mentioned Positive Aspects<br>• Smooth and Precise Movement: Many noted the arm moves fluidly, handling chess pieces accurately.<br>• Clear Voice/Explanation: Several answers highlighted that the robot's speech is generally understandable and coherent.<br>• Advanced Technical Ability: Respondents frequently praised its strong board-detection and analytical capabilities.<br><br>Most Mentioned Negative Aspects<br>• Overly Technical/Complex Language: A common concern was that its explanations could be too advanced or confusing for some audiences.<br>• Slow or Lengthy Process: Multiple people felt the scanning or movement took too long.<br>• Robotic or Distracting Voice/Noise: The monotone sound and mechanical noise were cited as off-putting or not engaging. |
| Output of Scenario B | Most Mentioned Positive Aspects<br>• Smooth and precise movement: The robot accurately picks up and places pieces without disturbing others, and its motion control is generally fluid and stable.<br>• Clear explanations: Many appreciated that it articulates moves clearly, providing reasoning and context for each chess move.<br>• Effective teaching approach: Explaining why certain moves are played and discussing strategy was seen as a strong educational feature.<br><br>Most Mentioned Negative Aspects<br>• Robotic or unnatural voice: Several found the voice to be lacking a human touch or warmth, making it less engaging.<br>• Noisy or distracting movements: The mechanical sounds during operation were frequently cited as disruptive.<br>• Limited interactivity: Users noted a lack of two-way communication or the ability to ask questions, reducing the personalized learning experience. |
| Output of Scenario C | Most Mentioned Positive Aspects<br>• Smooth and precise movement (many note its fluid, accurate handling of chess pieces)<br>• Intelligent or knowledgeable (demonstrates strong chess skills and offers insightful commentary)<br>• Entertaining or interactive (engages the user through narration, |
| | voice, and conversation)<br><br>Most mentioned negative aspects<br>- Loud or noisy (the whirring and mechanical sounds are off-putting to some)<br>- Bulky or large (the size and industrial look make it less home-friendly)<br>- Slow or boring (some feel its pace and monotone delivery can be dull over time) |
| Prompt to ChatGPT o1 to analyze Q6 | Based on the given answers to the question: "Do you have any recommendations about how to improve the robot?", summarize the most mentioned five recommendations in the answers. Make concise bullet points. Answers: |
| Output of Scenario A | - More natural, varied, or human-like voice<br>  (e.g., less robotic, different voice options, more expressive tone)<br>- Simpler, age-appropriate explanations<br>  (e.g., less technical jargon, clearer language for children)<br>- Friendlier or more appealing design<br>  (e.g., adding a face or screen, making it look less mechanical and more welcoming)<br>- Reduced mechanical noise<br>  (e.g., quieter motors, less distracting sounds during movement)<br>- Faster or smoother operation<br>  (e.g., increased speed, more fluid arm movements) |
| Output of Scenario B | - More natural, simpler speech: Many respondents suggested using a less robotic voice and avoiding overly technical language, especially for beginners or children.<br>- Reduce mechanical noise: A common recommendation was to make the robot's arm and movements quieter so it's less distracting.<br>- Friendlier design and smaller size: Several mentioned the robot's industrial look, proposing a more approachable or compact form—e.g., adding a face, giving it a name, or shrinking its footprint.<br>- Greater interactivity and Q&A: Multiple answers emphasized an option for students to ask questions and receive explanations in real time.<br>- Clearer movement cues: Suggestions included using a laser pointer or other visual indicators to highlight the squares being discussed. |
| Output of Scenario C | - Reduce mechanical noise: Many participants found the whirring and buzzing of the robot's arm distracting and suggested quieter motors or motion.<br>- Use a more natural, expressive voice: A large number of people felt the robot's monotone voice should be improved, either by adding emotion, varying intonation, or offering different accents.<br>- Make the design more compact: Several responses pointed out |
| | that the robot is bulky or tall, recommending a smaller, sleeker form factor.<br>• Improve overall aesthetics/approachability: Suggestions ranged from adding a "face" or a friendlier exterior to changing color schemes and reducing exposed wires.<br>• Enhance movement fluidity: Smoother, faster, or less "jerky" movement was mentioned to make the robot feel more natural and engaging. |
| Prompt to ChatGPT o1 to analyze Q7 – Scenario A | Based on the given answers to the question: "Do you think that within 10 years, robots like this could serve as a replacement of existing screen-based computers to help understand technical topics such as computer vision? Please elaborate on why or why not.", summarize the most mentioned three concerns in the answers. Make concise bullet points (keywords and explanation in one sentence). Answers: |
| Output of Scenario A | • High cost and limited cost-effectiveness: Many responses cite that these robots would be too expensive or impractical to deploy at scale compared to existing screen-based tools.<br>• Concerns about acceptance and human interaction: Several people doubt whether users would find a robot's impersonal or purely mechanical explanations as engaging as human teachers or more familiar screen-based interfaces.<br>• Incomplete or restricted functionality: Respondents worry that a robot arm cannot cover the breadth of content or flexibility offered by computers, and may not fully replace the versatility (e.g., code viewing, detailed visuals) screens currently provide. |
| Prompt to ChatGPT o1 to analyze Q7 – Scenario B | Based on the given answers to the question: "Do you think that within 10 years, robots like this could serve as a replacement of existing screen-based computers to improve chess skills? Please elaborate on why or why not.", summarize the most mentioned three concerns in the answers. Make concise bullet points (keywords and explanation in one sentence). Answers: |
| Output of Scenario B | • High cost: Many respondents believe the robots would be too expensive for widespread adoption.<br>• Limited portability: Several people noted that these physical robots are bulkier and less convenient than screen-based computers.<br>• Reduced versatility: Others highlighted that a single-purpose chess robot cannot match the broader functionality and advanced analysis tools offered by standard computers. |
| Prompt to ChatGPT o1 to analyze Q7 – Scenario C | Based on the given answers to the question: "Do you think that within 10 years, robots like this could serve as a replacement of existing screen-based forms of entertainment such as television? Please elaborate on why or why not.", summarize the most mentioned three concerns in the answers. Make concise bullet points (keywords and explanation in one sentence). Answers: |
| Output of Scenario C | - Lack of variety and visual/storytelling depth – Many respondents feel a robot cannot replicate the diverse content and creative appeal (shows, movies, storylines) that TV provides.<br>- Preference for passive entertainment – A common concern is that people want to "sit and watch," whereas a robot requires more active engagement and isn't as effortless as simply turning on a screen.<br>- Insufficient human/emotional element – Several mention that robots lack the emotional connection or relatable human factor that makes television and film compelling. |
| Prompt to ChatGPT o1 and Gemini 2.0 Flash Think to classify answers of Q7 – Scenario A | Based on the given answers to the question: "Do you think that within 10 years, robots like this could serve as a replacement of existing screen-based computers to help understand technical topics such as computer vision? Please elaborate on why or why not", classify each answer to one of the three classes: [positive, neutral, negative] and add the category to each row. Add quotation marks if any answer is not in quotation marks. The answers are provided as a csv. |
| Prompt to ChatGPT o1 and Gemini 2.0 Flash Think to classify answers of Q7 – Scenario B | Based on the given answers to the question: "Do you think that within 10 years, robots like this could serve as a replacement of existing screen-based computers to improve chess skills? Please elaborate on why or why not", classify each answer to one of the three classes: [positive, neutral, negative] and add the category to each row. Add quotation marks if any answer is not in quotation marks. The answers are provided as a csv. |
| Prompt to ChatGPT o1 and Gemini 2.0 Flash Think to classify answers of Q7 – Scenario C | Based on the given answers to the question: "Do you think that within 10 years, robots like this could serve as a replacement of existing screen-based forms of entertainment such as television? Please elaborate on why or why not", classify each answer to one of the three classes: [positive, neutral, negative] and add the category to each row. Add quotation marks if any answer is not in quotation marks. The answers are provided as a csv. |